%% file: main.tex
\pdfoutput=1

\documentclass[11pt]{article}

\usepackage[preprint]{acl}

\usepackage{times}
\usepackage{latexsym}

\usepackage[T1]{fontenc}

\usepackage[utf8]{inputenc}

\usepackage{microtype}

\usepackage{inconsolata}

\usepackage{graphicx}

\title{Paired Completion: Flexible Quantification of Issue-framing at Scale with LLMs}

\author{Simon D. Angus \\
  Dept. of Economics \& SoDa Laboratories \\
  Monash Business School \\
  Monash University
  \texttt{simon.angus@monash.edu} \\\And
  Lachlan O'Neill \\
  SoDa Laboratories \\
  Monash Business School \\
  Monash University \\
  \texttt{lachlan.oneill@monash.edu} \\}

\usepackage{amsfonts}       
\usepackage{nicefrac}       
\usepackage{hyperref}       
\usepackage{url}            
\usepackage{booktabs}       
\usepackage{xcolor}         

\usepackage{amsmath}
\usepackage{amsthm}
\usepackage{amssymb}
\usepackage{algorithm}
\usepackage{algpseudocode}
\usepackage{multicol}
\usepackage{multirow}
\usepackage{array, longtable}
\usepackage[normalem]{ulem}
\usepackage{listings}

\newtheorem{definition}{Definition}
\newcommand{\diff}[3]{\ensuremath{\Delta(#1, #2, #3) = \text{lp}(#3|#1) - \text{lp}(#3|#2)}}
\newcommand{\logprob}[2]{\ensuremath{\text{lp}(#2|#1)}}

\let\oldtable\table
\let\endoldtable\endtable
\renewenvironment{table}[1][h!]
    {\oldtable[#1]\small\rmfamily}
    {\endoldtable}

\begin{document}
\maketitle

\begin{abstract}
Detecting issue framing in text - how different perspectives approach the same topic - is valuable for social science and policy analysis, yet challenging for automated methods due to subtle linguistic differences. We introduce `paired completion', a novel approach using LLM next-token log probabilities to detect contrasting frames using minimal examples. Through extensive evaluation across synthetic datasets and a human-labeled corpus, we demonstrate that paired completion is a cost-efficient, low-bias alternative to both prompt-based and embedding-based methods, offering a scalable solution for analyzing issue framing in large text collections, especially suited to low-resource settings.
\end{abstract}

\section{Introduction}


It is widely held that public narratives have the power -- for better and worse -- to shape society~\cite{shiller2019narrative,patterson1998narrative,Graber2002MassMedia,BarabasJerit2009}. For quantitative social scientists, a typical analytical strategy to quantify the occurrence, characteristics and dynamics of these important narratives is to use a `framing' lens. According to the much-cited definition found in Entman \shortcite{Entman1993}, framing is the process by which individuals ``\emph{select} some aspects of a perceived reality and make them \emph{more salient} in a communicating text'' with the purpose of \emph{promoting} a particular interpretation or evaluation of reality. In essence, to frame, is to impose a world-view or `way-of-thinking' in communication, with the hope that others will be persuaded to be convinced of the same. In the standard approach to framing analysis~\cite{Chong2007}, one begins first, by identifying an \emph{issue} (e.g. `climate change'); second, by defining the \emph{dimensions} of that issue (e.g. `causes', `economic impact', etc.); third, by developing \emph{framings} of those dimensions (e.g. climate change/ causes/ framing: `anthropogenic emissions are responsible for climate change'); and then finally, by the manual labelling of texts (sentences, paragraphs) as to their framing alignment. Computational approaches to framing quantification have addressed various aspects of this decomposition, often (unhelpfully) under the generic heading of `framing' (we return to this point below)~\cite{Ali2022}.

Almost all prior approaches to computational framing analysis consider the task as a supervised machine-learning problem, typically as a multi-class classification task~\cite{Field2018}, and most often focusing on automatic labelling of \emph{dimensions} (e.g. `2nd Amendment', `Politics', `Public Opinion') of a single \emph{issue} (e.g. gun violence)~\cite{Liu2019,Zhang2023}, rather the more elaborate `world-view' like conceptualisation that Entman \shortcite{Entman1993} and Chong and Druckman \shortcite{Chong2007} hold. Where studies consider conceptual framing identification, large amounts of labelled data are required, and reported accuracy is modest (below 0.6)~\cite{Morstatter2018,Mendelsohn2021}, demonstrating the severe challenges inherent in automating an already difficult human-level task.

\begin{figure}[!thb]
    \includegraphics[width=\linewidth]{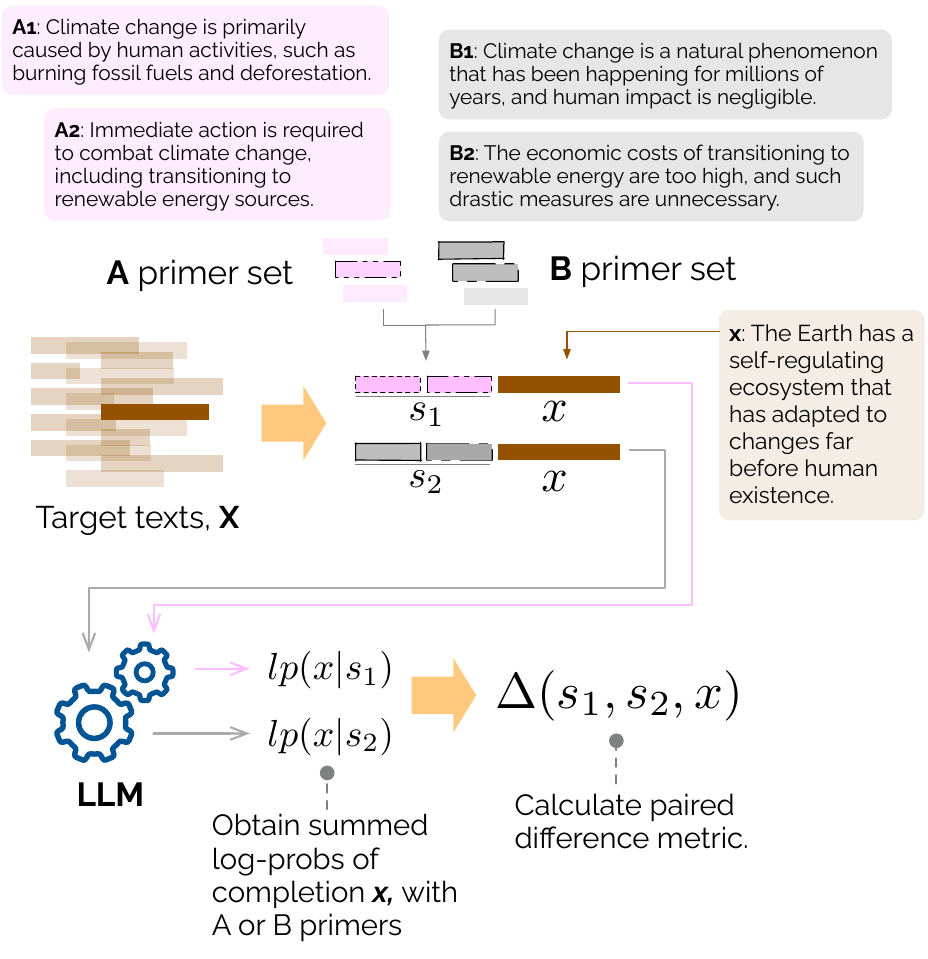}
    \caption{{\bf Paired Completion} -- Target texts ($X$) are taken, one at a time, as \emph{completions} to one ($k=1$) or two ($k=2$) priming conditioner sentences from two opposing issue framing sets, $A$, $B$, in turn. Summed log-probabilities of each completion text ($x$) are obtained from the LLM \emph{as if} the LLM had used the text to follow the conditioners. The two resultant summed log-probs inform the $Delta$ metric.}
    \label{fig:method-llm-logprobs}
\end{figure}

In this study, we introduce {\bf paired completion} -- a low-resource, `few-label', computationally efficient method that can accurately identify whether a target text aligns with one or other conceptual framing on a given issue (see Fig.~\ref{fig:method-llm-logprobs}). Importantly, and distinguishing it from previous methods, our approach: requires only a few (e.g. 5-10) example texts of a given framing (in fact, these can be generatively created); is low-bias compared to generative (prompt-based) LLM approaches; is significantly cheaper than generative approaches; and is highly flexible, switching issues or dimensions or framings is trivial.

Paired completion takes advantage of the log-probability (logprob) outputs of an LLM\footnote{Note: logprobs are available as outputs through the OpenAI API for ``babbage-002'' and ``davinci-002'' \citep{davinci-babbage-002}, and can be gathered by running the ``vLLM OpenAI-compatible API'' \citep{kwon2023efficient} on a local machine, for a wide variety of open-source models.} to find conditional probabilities of a text given a series of conditioners from different conditioning sets. We use the relative differences in probabilities to establish a baseline metric that (at least theoretically) is resilient to the model's prior probabilities of both the conditioning text and the text being aligned to the conditioning sets. We demonstrate empirically that this method is successful, and that one achieves superior performance from using this method with raw base models compared to ``asking'' instruct-fine-tuned AI-models the ``question'' at hand.

We conduct rigorous evaluation of our proposed method across a large synthetic dataset and human-labelled Immigration Tweet datasets. Our first study on synthetic data includes 192 independent experiments which compare paired-completion to four framing classification approaches over four diverse textual datasets, including two baseline approaches (traditional tf-idf vectors \citep{sparck1972statistical,salton1983introduction}, and fasttext sentence embeddings \citep{bojanowski2017enriching}) and three LLM-based methods (contextual embeddings \citep{peters2018deep,devlin2018bert,new-embedding-models} LLM chat token probabilities \citep{radford2019language}, and our novel \emph{paired completion} method). We demonstrate that the LLM-based approaches are, in general, far superior to the alternatives. The LLM-embedding approach is powerful with enough training data, but with small amounts of data (e.g. five sentences for each conditioning set) the LLM methods easily outperform LLM-embeddings. We also demonstrate that paired completion with LLMs is generally superior to the LLM prompting approach. We discuss why this might be the case in Section \ref{sec:diff}, from a theoretical perspective. We also conduct cost- and bias- comparison analysese at current gated API pricing to assess any trade-offs in performance. Our Immigration Tweet evaluation further demonstrates that paired-completion is strongly performant in diverse settings, especially when framing pairs under study are semantically contrasting.

\subsection{Contributions}
We introduce paired completion as a promising tool in the textual alignment task, especially in low-resource settings. We construct and make available a series of high-quality synthetic datasets. On both synthetic and real datasets we demonstrate that paired completion is a novel, efficient, low-bias alternative to either a chat-based LLM baseline or an embedding-training approach.

\subsection{Related Literature}

\subsubsection{`Framing' analysis}
Unfortunately, `framing' analysis does not have a clear definition in the computational literature, as evidenced by the variety of tasks that arise in a recent survey of 37 `framing' studies~\cite{Ali2022}. A starting point for many framing approaches is to leverage existing corpora of labelled datasets. Here, the media frames corpus (MFC)~\cite{card-etal-2015-media} and the gun violence frame corpus (GVFC)~\cite{Liu2019} have been the basis of many methodological contributions. However, these datasets conceptualise `framing' as \emph{dimensions} (ala Chong and Druckman~\shortcite{Chong2007}) of a topic or issue, not conceptual frames as we have distinguished earlier. The MFC is composed of 15 generic `frames'~\cite{boydstun2013identifying} such as `economic', `public opinion' and `cultural identity' applied to three issues (`immigration', `smoking', and `same-sex marriage'). Thousands of annotations were recorded as to whether one of these dimensions were associated with the text on a given issue. Likewise, the GVFC follows a similar approach, albeit tied more tightly to gun violence, `frames' are equivalent to \emph{dimensions}, and include `2nd Amendment', `Politics', and `Public Opinion'. Typically, computational methods approach framing in this way as a multi-class classification problem, using supervised machine learning methodologies such as featuring engineering, classifier selection and k-fold evaluation~\cite{Field2018,Liu2019,akyurek-etal-2020-multi,Zhang2023}. Common to all of these approaches is the need for large amounts of labelled ground-truth data, and consequently, the outcome that methods are not generalisable beyond the topics under study.

Where `framing' is implemented in a closer way to the conceptual framing we address in this work, challenges remain around the need for large labelled datasets and the accuracy of the methods. Morstatter et al.~\shortcite{Morstatter2018} consider support for, or against, 10 framings related to Balistic Missile Defence (BMD) in Europe over 823 online news articles (31k sentences). By writing out support- and opposition- (polarity) perspectives for each of the 10 framings, they are able to generate 20 framing-polarity classes, and apply traditional NLP methods to multi-class prediction. Alternatively, Mendelsohn et al.~\shortcite{Mendelsohn2021} label 3.6k social media posts (tweets) related to immigration for the 15 generic `framings' of Boydstun et al.~\cite{boydstun2013identifying} together with 11 conceptual framings (e.g. `hero', `victim', `threat' positions on immigrants). Using a base and fine-tuned encoder-only transformer approach~\cite{devlin2018bert,liu_roberta_2019}, they again conduct a multi-label classification study (we return to this dataset below). Whilst these examples are closely aligned to the same conceptual framing identification problem we address in the current study, each requires thousands of hand-labelled data to develop features to train traditional supervised machine-learning algorithms. Furthermore, and underlining the challenge of this task for traditional (even transformer based encoder methods), accuracy scores across 20 polarity classes in \citeauthor{Morstatter2018} and f1 scores across 11 conceptual framings in \citeauthor{Mendelsohn2021} are up to just 0.434 and 0.552 respectively.

Alternatively, and closer to the spirit of the present work, \citeauthor{Guo2022} \shortcite{Guo2022} take on a related but distinct task of quantifying the similarity between news sources by fine-tuning LLMs to each source and then conducting differential experiments on the likelihoods of word-completions when masking specific words within sentences on specific topics.


\subsubsection{Frame discovery}
A closely related task to textual alignment task is framing \emph{discovery}. Here, working in an unsupervised or semi-supervised manner, the task is to identify prominent framings in a corpus of texts. Examples include \citet{demszky-etal-2019-analyzing} who cluster tweet embeddings, \citet{roy-goldwasser-2020-weakly} and \citet{roy-etal-2021-identifying} who develop lexicons, and train embeddings to develop a frame identification model. The approach introduced in the present work complements these more elaborate discovery methods, especially in low-resource settings where advocacy groups are already aware of the `unhelpful' and `helpful' frames related to their domain, and so can move directly to known-framing detection.

\subsubsection{Stance detection}
A related task is that of stance detection, which we address briefly. Stance detection is typically formalised as the, ``automatic classification of the stance of the producer of a piece of text, towards a target, into one of three classes: {Favor, Against, Neither}.''~\cite{Kucuk2020} In effect, stance detection is a sub-problem of sentiment analysis, and again, typical approaches leverage traditional NLP techniques with labelled data as inputs~\cite{Kucuk2020}. Whilst paired completion shares the notion of `target texts', these texts work in concert to mark out a complex, nuanced conceptual framing on any issue, and the two priming sets (A,B) need not be strictly in opposition, but represent two perspectives, opening up more complex analytical insights than simply `favor' or `against'. 

\subsubsection{Perplexity}
One common measure of the capability of an LLM is perplexity \citep{jelinek1977perplexity}, which is a statistical measure of the model's ``surprise'' at a given completion under the logic that a model which is less surprised by correct answers is better (similar to the maximum likelihood principle). The paired completion approach developed in this work is a measure similar to perplexity, but instead of seeking the estimated likelihood of a particular completion we instead calculate and compare the likelihoods of multiple completions of a given text.

\section{Textual Alignment \& Paired Completion}\label{sec:definitions}


To hone in on \citeauthor{Entman1993}'s classic definition of framing, we reconceptualise the problem as one of ``textual alignment''. Namely, two texts on some topic or issue arise from the same conceptual framing, if they share a high level of \emph{textual alignment} -- a measure of the likelihood (in some sense) that the two texts might be spoken by the same entity (with a constant conceptual framing). This implies the statements come from the same theoretical outlook, model of the world, and/or causal structure. It is important that the expressive entity is generally defined. For we will be, at times, leveraging generative AI LLMs to play the role of $\mathfrak{E}$, alongside human expression, to quantify the degree of alignment.

\begin{definition}[Textual alignment]
    Given two conditioning texts $a$ and $b$, and an expressive entity, $\mathfrak{E}$ (e.g. a person, a generative AI LLM), text $x$ is said to be more textually aligned with $a$ versus $b$ if it is more likely that $x$ would be expressed by some $\mathfrak{E}'$ who previously expressed $a$, than the alternate case where $\mathfrak{E}'$ had previously expressed $b$.\label{def:alignment}
\end{definition}

Importantly, Def~\ref{def:alignment} is not the same as \emph{similarity}. Consider the texts, `Getting a dog will improve your life' and, `Getting a dog will ruin your life'. Whilst these are very similar (in fact, an LLM-powered contextual similarity score would be close to 1 for these texts), they are not \emph{textually aligned}. If someone holds the view that dogs \emph{improve} your life (framing A), it is highly unlikely that they would say that dogs \emph{ruin} your life (framing B). Yet these texts are highly similar on sentiment (both are neutrally posed) and share an almost identical vocabulary. However, consider the third text, `Pets help to keep you fit and healthy'. It is clear that this text is strongly textually aligned with framing A, but strongly dis-aligned with framing B. Yet, this text is perfectly dissimilar in vocabulary, and is of middling similarity in an LLM-powered contextual embedding space. These examples demonstrate that issue-framing, formalised as \emph{textual alignment}, is both `simple' for a human to perceive, yet difficult for existing computational methods (based on similarity, sentiment, vocab, embeddings) to detect.

As such, we desire a new set of tools to \emph{quantify textual alignment}. We consider these tools in the context of the ``Issue-Framing'' task, where a user wishes to detect and quantify texts from a large corpus which share the same framing, via textual alignment. Suppose the user has a small set of texts which together lay out a given framing position A, as compared to an opposing framing position B with a similar number of texts. We then formalise this task as follows:

\begin{definition}[The Issue-Framing Task]
    Given a corpus of texts $X$ (target texts) and a set of priming (or framing) texts $S = \{A,B\}$ comprising texts which represent framing A and B, for each $x \in X$, quantify the textual alignment towards $A$ and $B$.\label{def:task}
\end{definition}

Naturally, the user could accomplish this task by reading every text in $X$ and marking (labelling) whether the text is textually aligned with the conditioning or framing texts from $A$ or $B$. However, the aim of our work is to develop methods that might reliably accomplish this task at scale in an automated manner.


\subsection{Paired Completion}
We propose the ``paired completion'' method as a solution for the textual-alignment definition given above. Figure~\ref{fig:method-llm-logprobs} gives an overview of its components. Given some set of target texts on a given topic we wish to analyse, and a small set of texts which provide frames for perspective A and B on a given topic (e.g. `get a dog' vs. `don't get a dog'), we construct a pair of prompt sequences, $s_1 + x$ and $s_2 + x$ to pass to a generative LLM. Each prompt sequence is composed of a random selection from one of the priming sets (e.g. $s_1$ `get a dog'), followed by the target text ($x$).

For example, a prompt sequence could be `[priming text from A, $s_1$] Owning a dog will improve your life. [target text, $x$] Dog owners have lower blood pressure and less stress in general.' A similar sequence would be created for the same text $x$ with priming text(s) from set B. Each prompt sequence is then passed, one at a time, to a generative LLM, and instead of seeking a completion (i.e. generating new tokens) from the LLM, we instead exploit many LLM's ability to provide log-probabilities (the log of the likelihood that the model would have chosen that token/word next) for each token passed to the language model \emph{as if it had generated this exact sequence of text}. By so doing, we generate two conditional log-probabilities, $lp(x|s_1)$ and $lp(x|s_2)$ (see details in sec~\ref{sec:diff}), the conditional log-probs of $x$ being the completion to the priming sequence $s_1$ and $s_2$ respectively.

In this way, we are leveraging the twin features of LLMs: first, that LLM attentional mechanisms are highly adept at representing the latent semantic state of a given text; and second, that LLMs have been trained to provide \emph{coherent} sequences of text (i.e. to avoid {\it non sequiturs}). Together, the priming sequence will set the LLM on a particular statistical trajectory to keep the framing state consistent, which implies that if $x$ is within this trajectory (i.e. $x$ is textually aligned with the priming state), the summed log-probabilities the LLM assigns to the words in $x$ will be high. Whereas, if $x$ appears to contradict or speak for a different framing than the priming sequence, the log-probabilities for the words in $x$ will be very low. It is this difference that we exploit by testing both priming sequences from A and B to then calculate the Diff metric.

Paired-completion leverages LLMs' deep contextual representation of human meaning to evaluate text likelihood given prior context. We hypothesize that base LLMs, without task-specific fine-tuning or RLHF moderation, are best suited for this method, as such adaptations may compromise their fundamental language modeling capabilities.


See the appendices for details of the implementation of this method in evaluation.

\subsection{The Diff Metric}\label{sec:diff}
To quantify textual alignment, we introduce the Diff metric which measures the relative likelihood of a text following from different conditioning priors. Given a target text $x$ and two conditioning texts $s_1$ and $s_2$, we compute:

\[
\diff{s_1}{s_2}{x}
\]

where $\logprob{s}{x}$ represents the conditional log-probability of text $x$ given conditioning text $s$. The metric captures whether text $x$ is more likely to follow $s_1$ or $s_2$ by comparing their respective conditional probabilities. A positive value indicates stronger alignment with $s_1$, while a negative value indicates stronger alignment with $s_2$. This approach is robust to the model's prior probabilities of both the conditioning texts and the target text, as it considers only the relative difference in conditional probabilities. Detailed mathematical derivations and implementation specifics can be found in the appendices.

\section{Evaluation Approach}

We compare the novel paired completion method with a total of four comparison approaches, representing a mix of traditional NLP and transformer-based LLM methods -- three use a trained logistic regression~\citep{hosmer2013applied} classifier over varying training sample sizes, either employing TF-IDF vectors~\cite{sparck1972statistical,salton1983introduction}, word embedding vectors~\cite{mikolov2013distributed}, or LLM contextual embeddings~\cite{new-embedding-models} to represent texts in high dimensional space (see the appendices for details); whilst one uses LLMs via a prompt-based approach (described below).

\subsection{LLM Prompting}
Starting with a corpus of texts to test, we construct a prompt with three components: 1) a static instructional component which provides the LLM with the task information; 2) a set of context texts that represent framing A and B to be tested ($A,B$); and 3) a single target text ($x$). Unlike in LLM paired completion, we do not require the LLM to provide log-probs for the input sequence, but instead, we obtain the log-probs of the first two tokens produced by the LLM in response to this prompt, i.e. the first two generated tokens. Note that, by virtue of the constraints in the prompt, these probabilities include the log-probs for both response A and B.
We extract the probability of the first token of the label assigned to A (e.g. `[{\bf equality}]' [1 token]), and B (e.g. `[{\bf mis}][og][yny]' [3 tokens]), respectively.
With this information we can both identify which set the LLM has assigned the text to (based on the higher probability of its tokens) and calculate the equivalent Diff metric, $\Delta(A,B,x)$.

We used a fixed prompt across all models, initially optimized for GPT-4 and GPT-3.5, then adapted for Mixtral and LLaMA-2. Starting with GPT-4 proved suboptimal, as its tolerance for prompt imperfections may have influenced our design choices. While this approach potentially favored OpenAI models, our results ultimately showed open-source models performing better on the target tasks. We used a single prompt across all models in our final experiments.

\subsection{Performance Analysis}
In terms of true-positives (TP), false positives (PF) and false negatives (FN), the F1 score is calculated as,
\[
    f1 = \frac{TP}{TP + \frac{1}{2}(FP+FN)} \,\, .
\]
The f1 score takes a value from 0 to 1, and will be equal to 1 when the method perfectly identifies all the `As' in the data, and does not mis-identify any `Bs' as `As'.

Confidence intervals (95\%) for f1 scores were either calculated directly from replicates, in the case of the logistic regression methods (TF-IDF, word- and LLM- embeddings), and using bootstrapping in the LLM API cases (100 replicates, 1000 samples).

\subsection{Dataset Formation}
Existing large framing datasets like the MFC~\cite{card-etal-2015-media} and GVFC~\cite{Liu2019} focus on labeling \emph{dimensions} of topics rather than conceptual framings, making them unsuitable for our task. We therefore developed a synthetic dataset generation pipeline to create balanced, controlled examples of contrasting framings across multiple topics to provide a rich evaluation setting. In a second study, we use a small, human-labelled dataset on Immigration Tweets~\cite{mendelsohn2021modeling} to study paired-completion in more realistic settings. Our synthetic approach uses a two-step hierarchical process: first generating seed perspectives on a given topic, then producing sentences that align with each perspective. The pipeline ensures balanced representation across linguistic features to avoid confounding effects.

While synthetic data has inherent limitations for evaluation, it provides key advantages: controlled variation of framing elements, balanced representation across perspectives, and mitigation of potential training data overlap with the LLMs being evaluated. We generated datasets for four topics of varying complexity: dog ownership (as a straightforward baseline), climate change, domestic violence, and misogyny (representing more nuanced, controversial issues). Detailed generation protocols and validation procedures are provided in the appendices.

\subsection{Summary of Synthetic Experiments}

Together, across the five methods, four topics, and related variants, 192 experiments were conducted, as summarised in Table~\ref{tab:methods_models}.

\begin{table}[!htb]
\centering
\begin{tabular}{@{}lcccc@{}}
\hline
Method            & Models & Topics & Variants & Total \\
\hline
LR:TF-IDF          & 1      & 4      & 6        & 24    \\
LR:FastText        & 1      & 4      & 6        & 24    \\
LR:LLM Embed.        & 2      & 4      & 6        & 48    \\
LLM Paired Compl.  & 4      & 4      & 2        & 32    \\
LLM Prompt.       & 4      & 4      & 4        & 64    \\
\hline
\textbf{TOTAL}    & \multicolumn{4}{r}{\textbf{192}} \\
\end{tabular}
\caption{{\bf Summary of Experiments} The same four topics were tested across all configurations (`dog-ownership', `climate-change', `domestic violence', `misogyny'). For each LR (Logistic Regression) style experiment, 6 different sub-set sizes were used ($n \in \{10,20,50,100,200,500\}$). For LLM Paired Completion two variants for the number of conditioners were used ($k \in \{1,2\}$). For LLM Prompting, 4 prompt variants were used (seeds, distilled, summary, zero-shot).}\label{tab:methods_models}
\end{table}

\subsection{The Immigration Tweet Dataset}
To evaluate the performance of paired-completion in a more realistic setting, we applied paired completion to the labelled dataset from Mendelsohn et. al \cite{mendelsohn2021modeling}\footnote{We thank the authors for providing this dataset.}, focusing on the ``dev'' dataset of $450$ high-quality labelled tweets across various categories. All tweets were consensus labelled by two trained annotators with one or more of 11 specific frames (e.g. `Humanitarian: Immigrants experience economic, social, and political suffering and hardships'), within three main frames (`threat', `hero', `victim'), with the source containing a description of each frame (see Appendix for details). The descriptions were passed to Claude 3.5 Sonnet to create exemplar tweets as conditioners (nb: not the tweets from the dataset). To mimic the contrasting textual-alignment task with the tweet dataset, we re-sample the dataset into 55 subsets, where each subset contained a unique combination of $A|B$ labels ($c(c-1)/2$, for $c=11$ categories). The largest so formed subset contained 134 tweets (A:Humanitarian---B:Public Order), whilst the smallest contained just 9 tweets (A:Global Economy---B:War).

\section{Results \&  Discussion}
\begin{figure*}[!thb]
    \centering
    \includegraphics[width=1\linewidth]{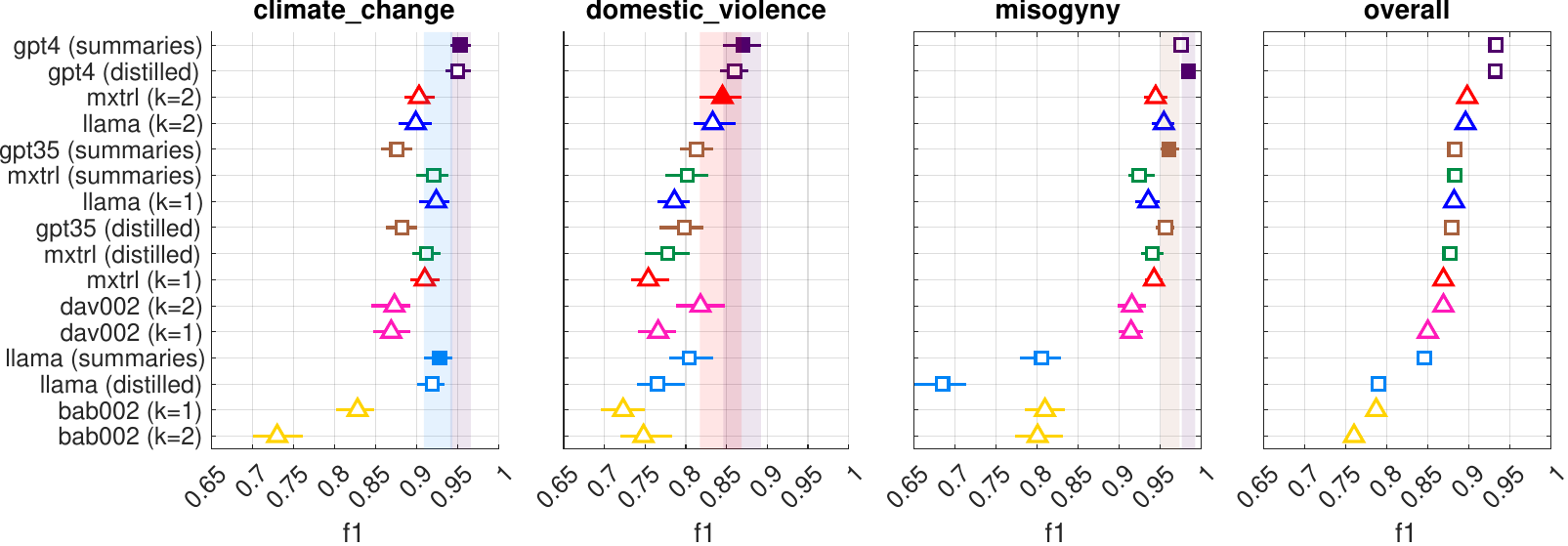}
    \caption{{\bf F1 outcomes across LLM prompting ($\square$) and paired-completion ($\triangle$).} Filled markers indicate approaches that are statistically similar to most performant method. Semi-transparent shading shows 95\% confidence interval for these methods to indicate other methods which provide similar performance to performant models. Ranking is by overall performance. See appendix for performance comparison with log-reg classification methods.}
    \label{fig:f1-concise-k12}
\end{figure*}

Our experiments demonstrate strong performance across the board for both prompt-based and paired completion methods, as shown in Figure \ref{fig:f1-concise-k12}. Paired completion methods tend to statistically perform the same or better than prompt-based methods. This section includes a broad summary of results. More detailed results, tables, and discussion can be found in the appendices.

\subsection{Comparative Analysis of Classification Methods}

With sufficient data (200+ samples), the embedding approach was competitive with GPT-4 prompting. However, embeddings performed significantly worse in few-shot learning contexts. Among LLM instruct models, GPT-4-Turbo outperformed all other models. GPT-3.5-Turbo, Mixtral-8x7b-Instruct-v0.1, and LLaMA-2-70B-Chat had similar performance, with LLaMA-2-70B-Chat having the highest propensity for failure modes. For the paired completion approach, performance trended with model parameter count, with LLaMA-2-70B performing best, followed by Mixtral-8x7b, davinci-002, and babbage-002. This consistency may occur because paired completion is less sensitive to model-specific factors like architecture, alignment, and fine-tuning.

\subsection{Cost vs. Performance}

\begin{figure}[!thb]
    \centering
    \includegraphics[width=\linewidth]{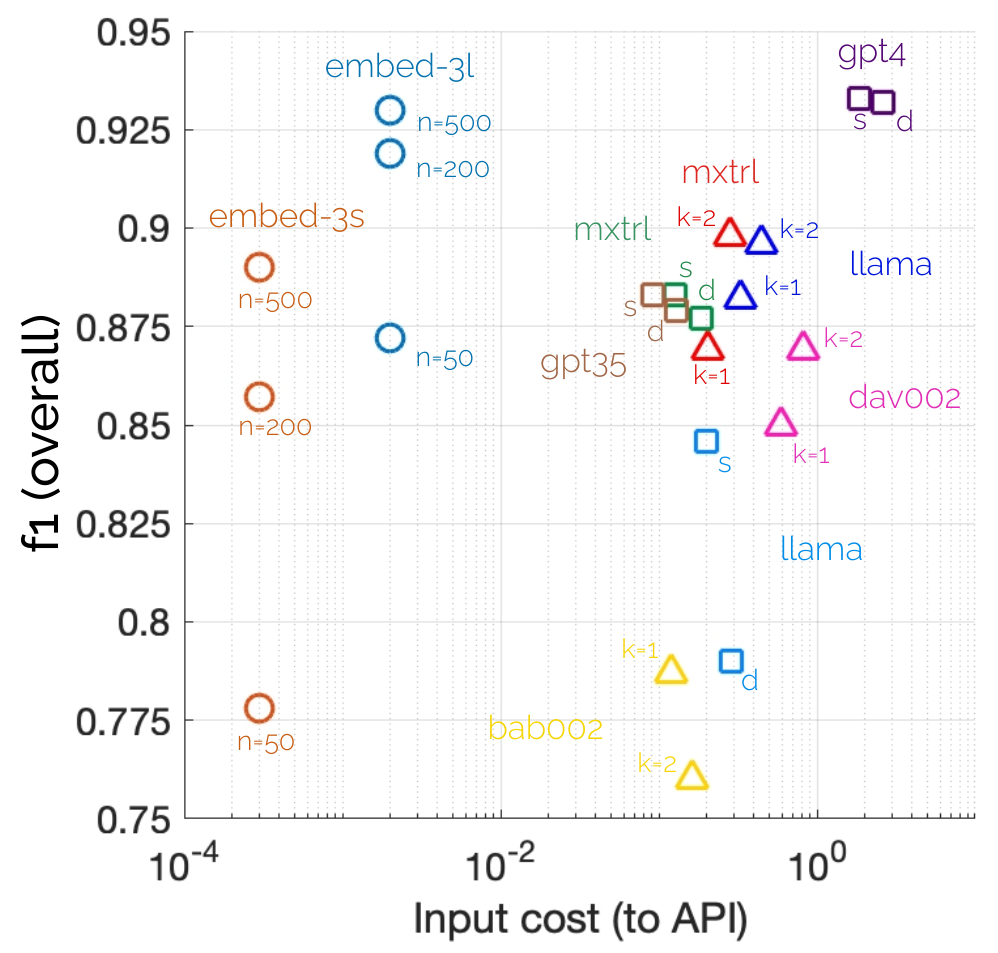}
    \caption{{\bf Cost --- performance trade-off for LLM methods.} Colouring and styling follows Fig~\ref{fig:f1-concise-k12}. Model short name and variant are provided for clarity.}\label{fig:cost_perf}
\end{figure}

An analysis of the cost-performance trade-off for the LLM methods (Figure \ref{fig:cost_perf}) reveals that the paired completion approach with LLaMA-2-70b and Mixtral-8x7b is very cost-effective for their level of performance. While GPT-4 had the best overall performance, it was also by far the most expensive. Other configurations can be chosen based on requirements and funding availability. All LLM-based approaches were significantly more expensive than the embedding approaches, which require more data but proved competitive given sufficient training examples.

\subsection{Model Bias}

We observed differences in the bias displayed by models and techniques that were dataset-dependent (see Appendix for details). Embedding-based approaches appear most robust to bias, with no statistically significant bias found for any embedding configuration. LLM-based approaches demonstrated bias in some scenarios, with the $k=2$ paired completion configuration potentially reducing bias compared to $k=1$. The top performing LLM paired completion methods (mxtrl-k=2; llama-k=2) show significantly less bias than the top LLM prompting approaches, including GPT-4. Further studies are needed to examine the sources of these biases, such as bias in training data, language modeling, or alignment. However, the results suggest the stronger LLM paired completion methods (e.g. llama-k=2) achieve a balance of high accuracy and low bias.

\subsection{Immigration Tweets}

We next apply the most performant paired-completion configuration from the synthetic study to the Immigration Tweet dataset from \citet{mendelsohn2021modeling}. 

\begin{figure}[!thb]
    \centering
    \includegraphics[width=0.9\linewidth]{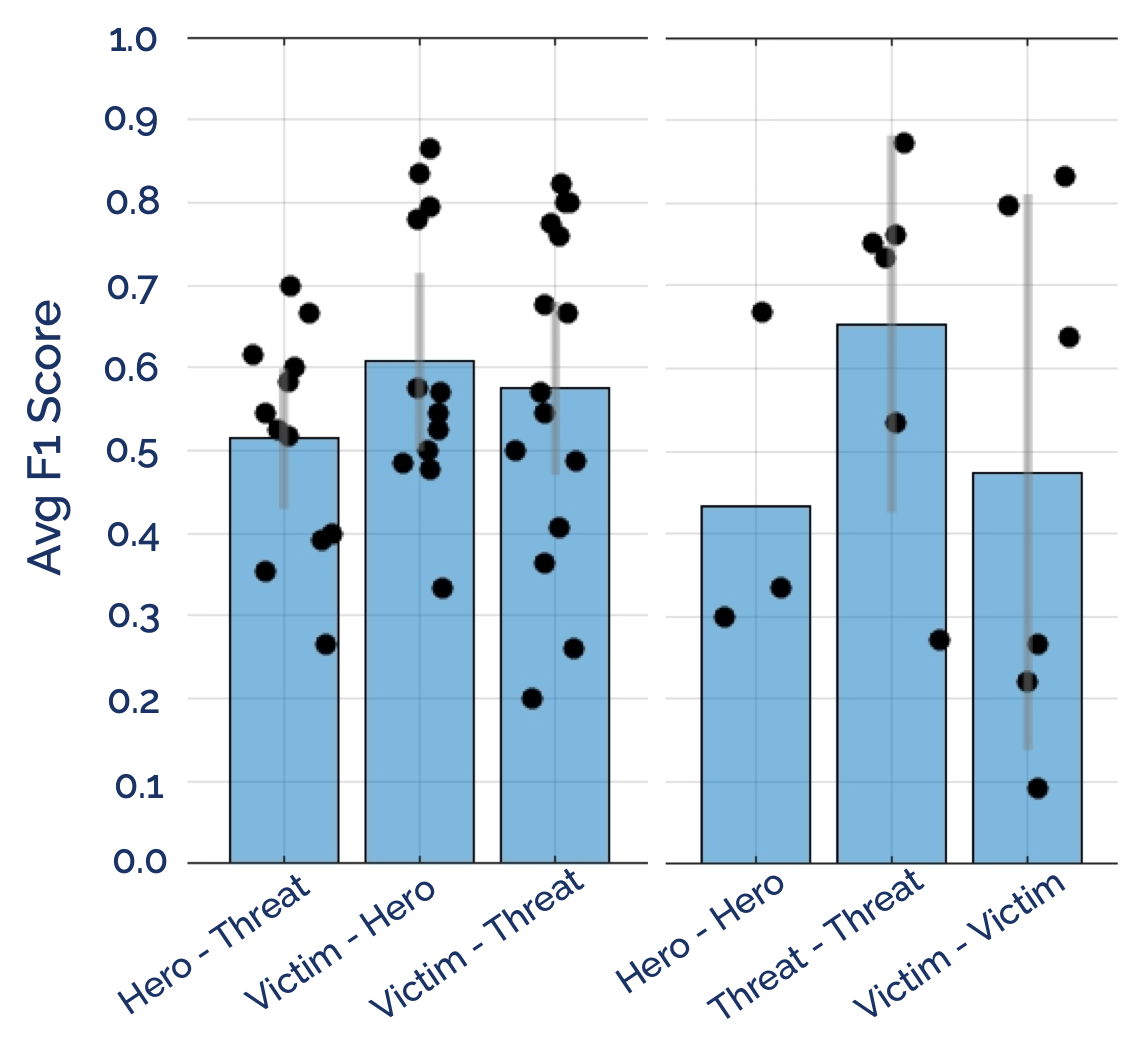}
    \caption{{\bf Immigration Tweet evaluation} Bars and error-bars indicate mean and 95\% CIs for F1 scores of all experiments within a main category pairing, whilst black markers indicate individual experiment means (x-axis jitter applied).}\label{fig:mendolsohn}
\end{figure}

In fig.~\ref{fig:mendolsohn} we present mean F1 scores across the 55 unique A-B experiments, grouped by main frame pairs. Averages for these groupings (reading L-R in the figure) were 0.51, 0.61, 0.57, 0.43, 0.65, 0.47 respectively. The highest F1 scores were obtained for the A:Public Order(threat)--B:National Cohesion(threat) (F1$=0.872$), A:Humanitarian(victim)--B:Cultural Diversity(hero) (F1$=0.864$), and A:Humanitarian(victim)--B:Integration(hero) (F1$=0.835$) pairs. In general, there is some indication that performance is stronger where the two frames are more distinct, i.e. taken from different main framing categories. For instance, the least peformant experiment was for A:Global Economy(victim)--B:Humanitarian(victim) (F1$=0.091$), with specific frame descriptions being barely distinct, A:`Immigrants are victims of global poverty, underdevelopment and inequality' vs B:`Immigrants experience economic, social, and political suffering and hardships'. Whilst direct comparison with the reference is not possible since they train a multi-stage multi-class labelling pipeline with thousands of labelled tweets, it is of note that the reported mean F1 score (across 11 specific categories) for the Issue-specific `dev' dataset is 0.550 (Table 3 in the reference), equivalent to our paired-completion findings obtained without training.

\section{Limitations \& Further Work}

\paragraph{Use of synthetic data for evaluation} - Whilst using synthetic data for evaluation has some benefits (described earlier) there are also some significant limitations. Principally, although our results with the small Immigration Tweets dataset are promising, ideally, a large expert annotation activity should be undertaken to generate a conceptual framing dataset covering a range of issues, and dimensions. Such a dataset would be of huge benefit to the field and would no doubt spur further refinement of framing analysis methods. As an intermediate step, a representative sample of our synthetic dataset could be validated by expert annotation to provide some comfort to our main findings.


\paragraph{Opposing framings, extension beyond binary classification}
The most performant evaluation results from the Immigration Tweets dataset back up our findings in the synthetic dataset, namely that paired completion is most effective in opposing framing contexts. However, it is possible that three or four distinct framings may also be successfully identified in corpora. The method's computational complexity scales linearly with the number of classes in terms of model calls, making it feasible for practical multi-class applications. Tentative experiments along these lines using the main frames in the immigration dataset suggest this is the case, but it remains for future work to explore the limits of paired completion in a many-frames setting.

\paragraph{Model bias in aligned models} -- The data seems to offer some support for the conjecture that aligned models are more prone to bias when performing framing alignment, but we cannot make any definitive claims without significantly more evidence and data. We only used three ``serious'' topics (climate change, domestic violence, and misogyny); for further study, we would significantly expand this (perhaps to 10, 20, or even 100 topics, ranging across and beyond, say, the Overton window \citep{russell2006introduction}).

\bibliography{custom,dem_resilience,nlp}

\appendix

\section{Note on Code \& Data Repository}

Code and synthetic data for replication are available at: \url{https://github.com/sodalabsio/paired-completion}. We do not expose the Immigration dataset as this was provided by the authors directly.

Refer to {\tt README.md} for instructions about code installation and replication.

Refer to the folder {\tt gpt-4-only-corpora/} to view the entire synthetic dataset.

\section{The Diff Metric}\label{sec:diff}
Suppose we have a set of $n$ priming sequences, $S = \{ s_1, s_2, ..., s_n \}$, and a set of $m$ target sequences $X = \{x_1, ..., x_m\}$. We wish to find the relative alignment, in some sense, of the elements within $X$ towards the different priming sequences in $S$.

We define the diff metric as follows:

\[
\diff{s_1}{s_2}{x}
\]

Note that $\Delta(s_1, s_2, x) = -\Delta(s_2, s_1, x)$.

The diff metric $\Delta$ describes the difference between the conditional probability of sentence $x$ after priming sequence $s_1$ and the conditional probability of sentence $x$ after priming sequence $s_2$. In practice, we calculate the prior probabilities of all priming sequences $s \in S$ as $p_s$, and all texts in $x \in X$ as $p_x$, and the probability of a concatenated string $s + x$ as $p_sx$. Note that concatenation is not necessarily simple string concatenation, but rather ensures grammatical correctness - there is no perfect way to do this, but we found that just ensuring grammatical correctness seems to work sufficiently well in practice.

We then compute $\logprob{s}{x} = p_{sx} - p_s$ to find the conditional probability of $x$. We can compare this to the prior probability $p_x$ to determine whether the presence of $s$ has made $x$ more or less likely, and we can compute $\logprob{s_1}{x} - \logprob{s_2}{x}$ (i.e. the $\Delta$ metric). Since a larger logprob indicates a higher probability, $\Delta$ will be positive if $x$ is more likely after $s_1$ than after $s_2$, and negative if $x$ is less likely after $s_1$ than after $s_2$. Because LLMs (and language models in general) might assign different prior probabilities to both the conditioning sentences $s$ and the alignment text $x$, any such method must be robust to priors. This is why we use the \textit{difference} in conditional probabilities of the same text with different prompts, which is robust to the prior probabilities of both $s$ and $x$.

One interpretation of this approach, with reference to Def.~\ref{def:alignment}, is that the LLM performs the role of the expressive entity $\mathfrak{E}$, and so provides a quantification of the likelihood that the text $x$ follows text $s_1$, versus following text $s_2$, i.e. we obtain a measure of \emph{textual alignment}.

Since the core idea of paired completion is to use the priming/conditioning sequence to statistically deflect the LLM towards the given framing (and so, measure the model's degree of `surprise' with the completion text) we conjecture that a longer conditioning sequence may lead to improved accuracy in classifying and retrieving texts aligned with a given framing. To explore this possibility we test two treatments, with either one ($k=1$) or two ($k=2$) priming/conditioning text(s) being used. Implementation details are provided in the appendices.

\section{Comparison methods used in evaluation: details}

As mentioned in the main paper, the {\bf paired completion} method is compared with two other broad classes of methods:
\begin{itemize}
    \item {\bf LLM prompting} -- where LLMs are engaged as human-like labellers, being tasked with making a labelling decision on a given target text relative to two alternative framings in context; or 
    \item {\bf Methods with labelled data} -- here we assume that a large amount of labelled data are available to train more traditional classifiers. We compare three variants of these, each using a logistic regression classifier with either tf-idf vectors, word-embedding vectors, or contexual embedding vectors from LLM encoder models.
\end{itemize}

We provide here additional information on each alternative method, introduced in the main paper.

\subsection{LLM Prompting}

 Instead of employing the LLM as an expressive entity which can also provide quantification to conditional text completions (as in LLM Paired Completion), we switch to using the LLM in its generative/chat mode, and ask it directly to assess, based on priming information, which framing a given text belongs to. Whilst this could be said to shift the task towards a `knowledge-model' capability of LLMs (i.e. leveraging its embedded \emph{knowledge} of the human world via training), by careful construction of the prompt, we can induce the model to operate closer to its `language-model' modality.

\begin{figure}[!htbp]
    \centering
    \includegraphics[width=\linewidth]{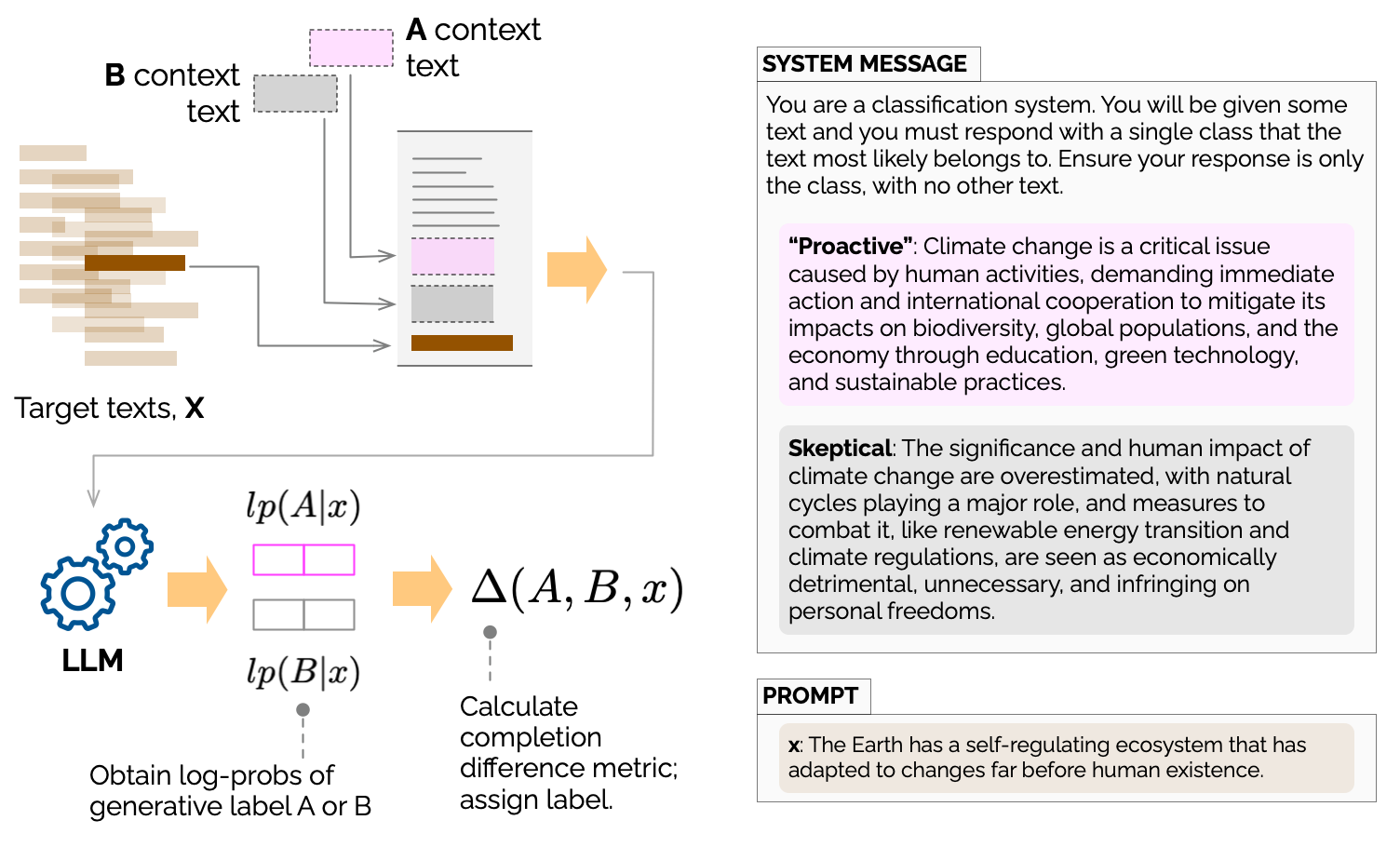}
    \caption{{\bf Narrative classification with the LLM Prompting method.} Texts (which for our experiments are labelled, but need not be so in use) to be classified are passed one at a time to a Chat- or Completion- based generative LLM within a prompt. The prompt is formed by concatenating together static task description text, followed by some priming text for framing A and framing B. We explore four different priming text approaches including: seeds, distilled, summaries and zero-shot (see text for details). The LLM task is to provide the most likely label for the given text ($x$) with log-probabilities (log-probs) obtained for the first tokens of each A/B label enabling the calculation of the $\Delta$ metric. An example system message and prompt for the Climate topic is shown at right, with \emph{summaries} shown for context texts A and B.}
    \label{fig:prompt}
\end{figure}

In Fig.~\ref{fig:prompt} we provide an overview of our method. Again, starting with a corpus of texts to test, we construct a prompt with three components: 1) a static instructional component which provides the LLM with the task information; 2) a set of context texts that represent framing A and B to be tested ($A,B$); and 3) a single target text ($x$). Unlike in LLM paired completion, we do not require the LLM to provide log-probs for the input sequence, but instead, we obtain the log-probs of the first two tokens produced by the LLM in response to this prompt, i.e. the first two generated tokens. Note that, by virtue of the constraints in the prompt, these probabilities include the log-probs for both response A and B.\footnote{Note: every token produced by an LLM is just one choice from a very large token vocabulary known to the LLM. The LLM's core inference task is to assign a probability to each token, surfacing the most likely token to the top which is then fed back into the LLM in an auto-regressive manner. Here, we extract the probability of the first token of the label assigned to A (e.g. `[{\bf equality}]' [1 token]), and B (e.g. `[{\bf mis}][og][yny]' [3 tokens]), respectively.} With this information we can both identify which set the LLM has assigned the text to (based on the higher probability of its tokens) and calculate the equivalent Diff metric, $\Delta(A,B,x)$.

We use a fixed prompt, which was initially fine-tuned for GPT-4 and GPT-3.5, and then further tuned for Mixtral-8x7b-Instruct-v0.1 and LLaMA-2-70B-Chat. It is possible this somewhat biased the prompts towards the OpenAI models, but this is difficult to mitigate in general. In hindsight, it was a mistake to tune our prompts for GPT-4 first, as while GPT-4 was almost certainly going to give the best performance on the tasks at hand (compared to the other models in consideration), it was also a lot more forgiving of errors, confusing wording, and conflicting instructions within the prompt. We conjecture (but do not know) that tuning prompts with weaker models first might lead to better overall results, exactly because these models are less forgiving of such mistakes.

We found that Mixtral was a reasonably straightforward (almost drop-in) replacement for GPT-3.5, while the LLaMA model's output format included an extra space token before the output. We do not know whether this is due to a formatting issue on our part, a quirk of LLaMA-2, or something else, and we thus cannot exclude the possibility that performance has been left on the table for LLaMA-2. However, given the prevalence of the OpenAI models, and the relative ease of applying the Mixtral model, we feel it is not unreasonable to expect other LLMs to adopt to industry norms (especially when presented through a drop-in replacement for the OpenAI API, as both these LLMs were when used with chat prompting).

When crafting the prompts, a decision must be made about how to represent the frames A and B in the prompt. This decision leads to a trade-off between cost and performance. LLM classification is theoretically quite cheap as it only requires the input text and one output token, and the most expensive part of the LLM call (from a token budget perspective) is the set of instructions, which includes an explanation of the problem, the expected output format, and a list of classes. Hence, providing the model with more text to represent a given frame may induce higher accuracy but at the cost of more input tokens, and so, overall task cost on pay-for-inference LLM endpoints.

To explore this trade-off, we developed four approaches to frame representation of varying levels of detail (a complete example (the Dog Ownership dataset can be found in Section~\ref{dog-ownership}) as follows:
\begin{enumerate}
    \item{\bf Seeds} The full list of `seeds' used to generate the dataset. A single seed is a single sentence consistent with the given frame. For example, the climate change dataset contains 20 seeds for each of the two classes `science' and `denialism'.
    \item{\bf Distilled} A distillation of the seeds of each class into five pairs per class.
    \item{\bf Summarized} A single sentence for each of the two classes providing an overall summary of the seeds for each class.
    \item{\bf Zero-Shot} Only provide the names of each class.
\end{enumerate}

In the main paper, we focus on results from only the `distilled' and `summary' versions of the prompt since these are closest in equivalence to the LLM Paired Completion method which uses one ($k=1$) or two ($k=2$) distilled sentences only in the conditional sequence. Zero-shot is included as a `raw' test of the LLM's capabilities and shifts the task closest to a `knowledge-model' modality since no examples are provided to the model at all of each frame, simply the frame names. Full results that include `seeds' and `zero-shot' are included in Table~\ref{tab:f1-llms} below.

\subsection{Methods with Labelled Data}

With synthetic labelled data in hand we first apply a standard train/test split, reserving 500 samples for training and 300 samples for testing. To explore the role that training data counts have on performance, we undertake six independent experiments for each labelled data method, taking a random sub-sample from the training examples of size $n=10$ up to size $n=500$.

\begin{figure}[!htbp]
    \centering
    \includegraphics[width=\linewidth]{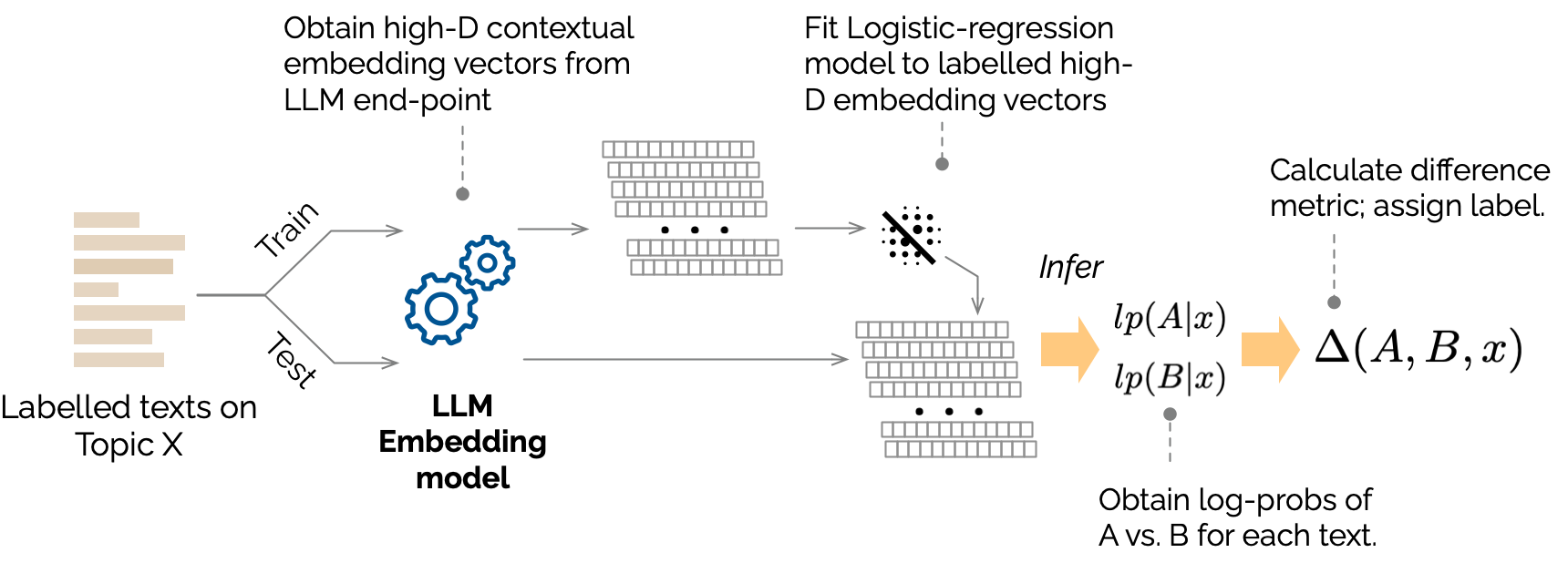}
    \caption{{\bf Narrative classification with LLM embeddings.} Labelled texts are first split into train/test sets before obtaining LLM (encoder-only) contextual embedding vectors (e.g. length 1536 (OpenAI's ``text-embedding-3-small''), or length 3072 (OpenAI's ``text-embedding-3-large'') for each text. A penalised logistic regression model is fitted to the labelled vector data and then applied to the test vectors to obtain accuracy and $\Delta$ metrics for analysis. Note: LLM \emph{contextual} embeddings should not be confused with earlier \emph{word}-embedding models, such as `FastText`~\cite{mikolov2013distributed} used in our comparison experiments.}
    \label{fig:embed}
\end{figure}

For each method, we use the same (standard) NLP machine learning approach to classification by training a (penalised) logistic regression model on vector representations of each text following defaults in the ``sci-kit learn'' package~\cite{scikit-learn}. We form vector representations under three scenarios as follows. This approach enables us to obtain log probabilities for each class, and so, the $\Delta$ metric as before (refer Fig.~\ref{fig:embed}).

The three labelled approaches are as follows.

\subsubsection{TF-IDF}
Term Frequency -- Inverse Document Frequency (TF-IDF) vectors aim to represent a text in a corpus by computing a vector of fixed length corresponding to the term vocabulary of the corpus, each element of which is built by multiplying the frequency of the given term in the text, by the inverse of the frequency of the term across all texts (documents) in the corpus. In this way, high values in the TF-IDF vector for a text will be given to frequent terms in the text which are also relatively rare in the corpus.

\subsubsection{FastText Word Embeddings}
Alternatively, we generate a pre-trained word embedding for each term in the text using the FastText word embedding model~\cite{bojanowski2017enriching}. These models are trained on large amounts of text, using the context of each word to infer the semantic position in a high-dimensional space of a given word. Words which are used in similar contexts thus tend to be close to one another within such a space (e.g. `walk' and `run'). To create a single vector to represent the text, average word vectors are obtained.

\subsubsection{LLM Contextual Embeddings}
As opposed to traditional word embeddings like FastText, which embed individual words in a semantic vector space and then average the vectors to find an aggregate representative of a document, contextual embeddings such as OpenAI's~\cite{new-embedding-models} ``text-embedding-3-small'' and ``text-embedding-3-large'' use a LLM architecture, but output a representative (encoding) vector for an entire text rather than token completions. The major advance of this method is that the entire sentence/text is embedded at once by the model, as opposed to word-at-a-time and averaging in the word embedding approach. Unsurprisingly, given that such models are trained on a much larger corpus of training data, and with many more parameters, than word-based emebeddings, they have been found to out-perform traditional embedding approaches on standardised tasks.\footnote{See supremacy of LLM embedding models at, for example, Huggingface's MTEB leaderboard: \url{https://huggingface.co/spaces/mteb/leaderboard}.}

\section{Bias study: results detail}

\begin{figure}[!thb]
    \centering
    \includegraphics[width=\linewidth]{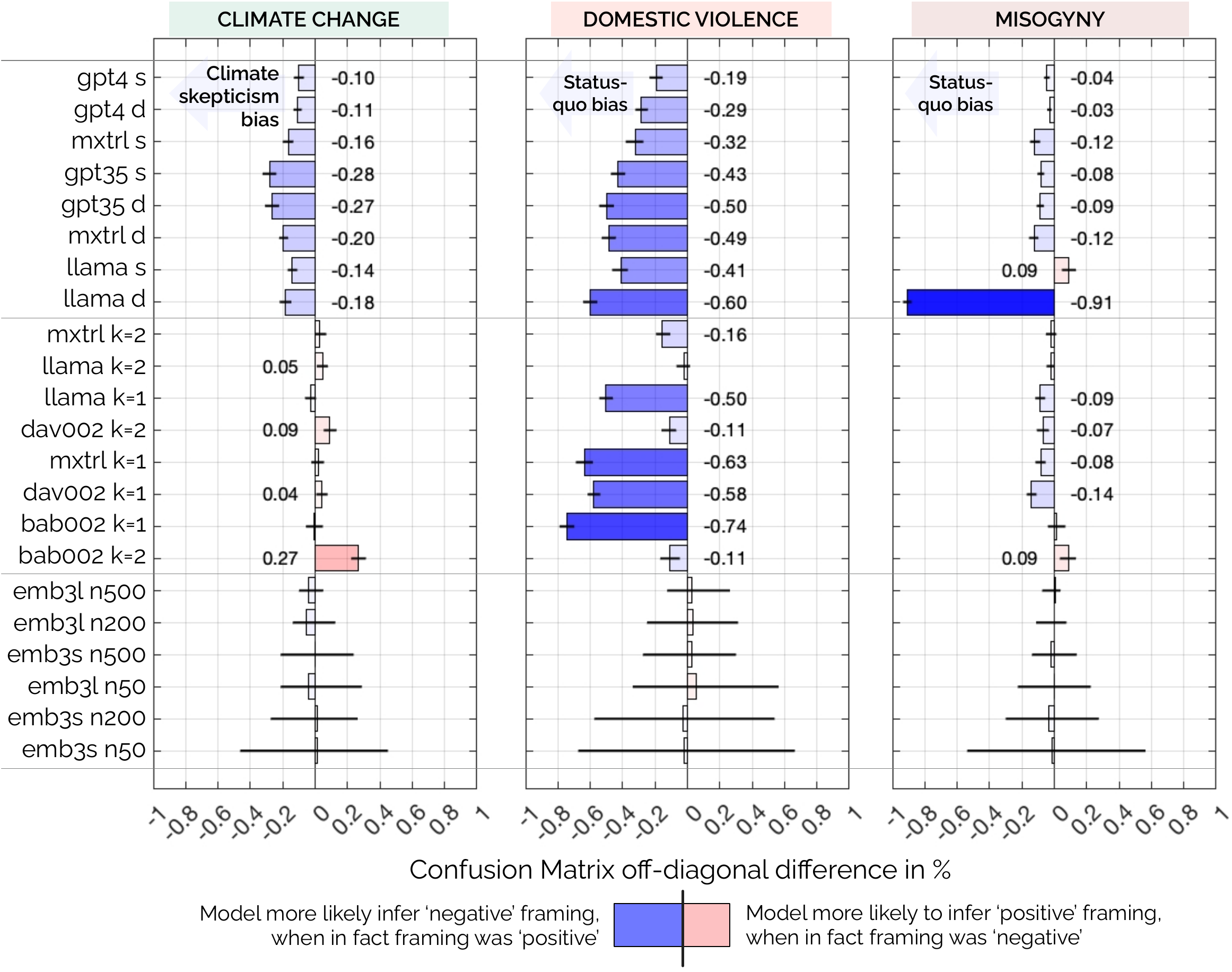}
    \caption{{\bf Inference Bias} Mean and (95\% confidence intervals) for issue-framing asymmetry, or model bias, calculated as the difference between the off-diagonals in a normalised confusion matrix. Scores are given where non-zero bias is statistically significant ($p < 0.05$).}\label{fig:model-bias}
\end{figure}

In figure~\ref{fig:model-bias} we present the details results of the bias study, referred to in the main paper, indicating the low relative bias of the most performant paired-completion approach, compared to all of the main prompt-based approches (top panel).

\section{Comparative Analysis \& Notes on Classification Methods}

In this section we provide tabular results and commentary to complement the main results figure of the main paper.

In Table~\ref{tab:f1-llms} we present f1 results for LLM based methods including paired-completion and LLM prompting. For the latter, we includethe  `seeds' and `zero-shot' variants of the LLM Prompt methods for comparison.
\input{figs/f1_table}

In Table~\ref{tab:f1-lab} we provide results tables for the labelled methods. In these experiments, each row of results represents a treatment where a given sub-set of the labelled training data was used for training the model. For example $n=100$ implies that 100 examples from the training data were exposed to the logistic classifier to fit the model. Note that the test set was held-out prior to any sampling and was consistent across methods.
\input{figs/f1_embedding}

\subsection{Traditional Classification}
Both the tf-idf and fasttext-based classification methods showed inferior performance to the LLM-methods. The tf-idf approach generally performed better than the fasttext approach, indicating that on these datasets a keyword approach is superior to a summed embedding vector approach. This does not include the contextual embeddings, which are more complex than the simple summation performed by fasttext when embedding a sentence. The superiority of LLM-based approaches was, of course, expected, and these traditional methods were included to provide a baseline for performance.

\subsection{Contextual Embeddings}
The significant improvement in the performance of the contextual embedding models, compared to the non-contextual fasttext embeddings, demonstrates the importance of contextuality when creating embedding vectors for text. We observed a significant uplift in the performance of contextual embeddings correlating with dataset size, far more than the uplift between any of the prompting methods (e.g. seeds vs summaries). With a large enough amount of data, generally 200 or more samples, the embedding approach approached or even exceeded the performance of GPT-4 with prompting, demonstrating the potential power of this approach. However, the embedding approaches did significantly worse in the few-shot learning contexts (e.g. with 10 samples per class, which is still double the number of examples provided to the LLMs with the distilled prompting approach).

We therefore conclude that contextual embeddings can be a good, and potentially cost-effective, method for performing classification in contexts with large amounts of training data, but they are not as suitable when there is little training data (i.e. in a few-shot learning context). A hybrid approach, where one generates a training corpus with LLM classification and then uses this to train an embedding system, might be cost-optimal, but analysis of this approach is beyond the scope of this study.

\subsection{LLM Instruct Models}

We observed general superiority from GPT-4-Turbo, the most powerful LLM model available (and the only model of its performance class to support outputting logprobs, making it suitable for classification in our logprob-based pipeline). This was expected, as none of the other models claim parity with GPT-4, and we therefore use GPT-4 as an upper bound on performance (similar to using traditional classification as a lower bound on performance).

The other LLM instruct-capable models, including OpenAI's GPT-3.5-Turbo, Mistral's Mixtral-8x7b-Instruct-v0.1, and Meta's LLaMA-2-70B-Chat, went blow for blow throughout the experiments, though LLaMA-2-70B-Chat demonstrated the highest propensity for failure modes. We recommend trying several models to determine which is most suitable. GPT-3.5-Turbo seems to be a relatively dependable choice, and the relative reliability of the OpenAI API coupled with relatively high rate limits make it a straightforward option for running large experiments. Mixtral-8x7B generally seems to be as good as, if not better than, LLaMA-2-70B-Chat, which is in line with previous experimental results \citep{jiang2024mixtral}.

\subsection{LLM Paired Completion}

The Paired Completion approach requires an API which supports, in OpenAI API parlance, ``echo[ed]'' logprobs (i.e. outputting the logprobs for input tokens). For unknown reasons, OpenAI launched and then subsequently disabled this feature on their ``gpt-3.5-turbo-instruct'' model, and to our knowledge have never offered it on their Chat API (only their ``legacy'' Completions API). However, they do support the feature on their legacy completion models, including ``davinci-002'' and ``babbage-002''. vLLM also supports the ``echo'' parameter through its OpenAI-compatible API, which we leveraged to get results from LLaMA-2-70B and Mixtral-8x7b-v0.1. Note that none of these models are fine-tuned for chat.

We found babbage-002 generally performed poorly compared to davinci-002, which in turn was outperformed moderately by the two open-source models. The performance trend was relatively stable, with LLaMA-2-70B performing best, Mixtral-8x7b close behind LLaMA, davinci-002 close behind Mixtral, and babbage-002 quite behind the pack.

We conjecture that this consistency in performance occurs because the paired completion approach is less sensitive to outside influences such as architectural changes that make model training easier (which are continually developed as the literature expands), alignment (via mechanisms such as RLHF), fine-tuning for instruct/chat, and the size of the datasets used for post-training tuning steps. It may also be that these problems (which were designed for use with davinci-002 and babbage-002) are too easy for the newer, more powerful models, and that more complex experiments would tease out more distinctions between them. It should be noted that the performance trend places the models in order of their number of parameters (although the parameter counts of davinci-002 and babbage-002 are only estimated, we suspect that the models are indeed placed in correlation with their parameter count).

\section{Computational Complexity and Resource Requirements: additional comments}
Both the paired completion and prompting approaches increase time complexity linearly with the number of classes. The paired completion is non-comparative, in that a new set of framings can be added independently of past/future framing sets, and thus the compute scales linearly with the number of framings (though it might scale faster than linearly with the number of framings \textit{within} the framing set if using $k>1$, as the number of comparisons for a framing set of size $n$ is $O(n^k)$).

The prompting approach only requires a single call to the model, regardless of how many classes are used, but the number of tokens used within the call will scale linearly with the number of classes. There will be a large constant term in the size of the input prompt corresponding to an explanation of the problem, the expected output, and the required output format, meaning that for smaller numbers of classes, the relative increase in prompt size can be small. However, the difficulty of the task also increases with the number of classes, and we conjecture that the paired completion approach will scale better to a very large number of classes, as the model only needs to ``consider'' one class (i.e. one priming sequence) at a time.

From a practical perspective, the OpenAI API has a seemingly little-known feature that allows calls to the Completions API to be batched (thus including multiple texts in a single API call, and receiving all the results for those texts in a single response to the API call). We found this a useful speed boost (by a factor of 20x) when using babbage-002 and davinci-002, because we were primarily rate-limited by API calls rather than tokens used. However, our experience is that other OpenAI-compatible vendors tend not to implement this feature, and it's unclear if there would be time savings from this anyway (as they might not execute the calls in parallel in their backend).

\section{Cost analysis: details}

We provide here, in Table~\ref{tab:cost-llms} tabular data for the cost-performance figure presented in the main paper, plus in Table~\ref{tab:cost-embed} the same for the embedding labelled data approaches which utilise the embedding end-point.
\input{figs/costs_table}

GPT-4 proved by far the most expensive model, which was the expected result and the price (in a quite literal sense) paid for its excellent performance on all benchmarks. Other models varied in cost, but as demonstrated in the main paper, the paired completion approach with LLaMA-2-70b and Mixtral-8x7b proved very cost-effective for their performance. Trade-offs can be made based on requirements and funding availablility, but all LLM-based approaches were significantly more expensive than the embedding approaches. These require sufficient data (up to two orders of magnitude more than the LLM approaches), but proved competitive and cost-effective given enough data.

\clearpage
\newpage
\section{Algorithms}\label{app:algorithms}
\subsection{Stratified N-Tuple Sampling Algorithm}

This algorithm is used in paired-completion to ensure that priming (conditioning) texts are sampled equally and representatively across all text log-prob inference calls.

This algorithm accepts a set of $values$ and returns a set of the same size, where each element is a tuple of $sample\_size$ values from the $values$ set, there are no duplicates in a tuple, and each value is used in the same number of tuples.

\begin{algorithm*}
\caption{Stratified N-Tuple Sampling}
\begin{algorithmic}[1]
\Procedure{StratifiedNTupleSampling}{$values$, $sample\_size$}
    \State $stack \gets []$
    \State $counts \gets \{\}$
    \For{$value$ in $values$}
        \State $stack.extend([value] * sample\_size)$
        \State $counts[value] \gets sample\_size$
    \EndFor
    \State $random.shuffle(stack)$
    \State \textbf{assert} $(len(stack) == len(values) * sample\_size)$
    \State $batches \gets []$
    \For{$\_$ in $range(len(values))$}
        \State $batch \gets []$
        \For{$\_$ in $range(sample\_size)$}
            \State $idx \gets 0$
            \State $max\_count \gets max(counts.values())$
            \While{$stack[idx]$ in $batch$ or $counts[stack[idx]] \neq max\_count$}
                \State $idx \gets idx + 1$
            \EndWhile
            \State $val \gets stack.pop(idx)$
            \State $batch.append(val)$
            \State $counts[val] \gets counts[val] - 1$
        \EndFor
        \State $batches.append(batch)$
    \EndFor
    \State \textbf{assert} $(len(stack) == 0)$
    \For{$value$ in $counts$}
        \State \textbf{assert} $(counts[value] == 0)$
    \EndFor
    \State \Return $batches$
\EndProcedure
\end{algorithmic}
\end{algorithm*}

\subsection{Dataset Generation Algorithms}
Here we include the algorithms used for dataset generation. System messages and other prompts used to generate the datasets are included in the next section.

\begin{algorithm*}
\caption{Generation methods for seeds and sentences.}
\begin{algorithmic}[1]
\State \textbf{Input:} Topic $T$, Number of seeds $N$, Sentences per perspective $K$
\State \textbf{Output:} Synthetic dataset $D$ with structured perspectives and sentences

\Procedure{GenerateSeeds}{$T, N$}
    \State Use a language model to generate $N$ seed perspective pairs on topic $T$
    \State Each seed pair is structured as $\{a: \text{"perspective A"}, b: \text{"perspective B"}\}$
    \State \Return List of seed perspective pairs
\EndProcedure

\Procedure{GenerateSentences}{$\text{Perspective}, K$}
    \State For each perspective, generate $K$ sentences using a language model
    \State Each sentence reflects the given perspective's viewpoint
    \State The sentences for each perspective are structured as a list: $\{1: \text{"sentence 1"}, 2: \text{"sentence 2"}, ..., K: \text{"sentence K"}\}$
    \State \Return List of sentences for each perspective
\EndProcedure

\end{algorithmic}
\end{algorithm*}

\begin{algorithm*}
\caption{Generation methods for distilled, summarized, and named prompts}
\begin{algorithmic}[1]
\Procedure{DistilSeeds}{$\text{Seeds}$}
    \State Input the seeds to a language model
    \State Distil the seeds into a smaller number of representative pairs
    \State Each distilled pair maintains the structure $\{a: \text{"distilled perspective A"}, b: \text{"distilled perspective B"}\}$
    \State \Return List of distilled perspective pairs
\EndProcedure

\Procedure{SummarizeSeeds}{$\text{Seeds}$}
    \State Input the seeds to a language model
    \State Summarize the seeds into a single perspective pair
    \State The summarized pair is structured as $\{a: \text{"summarized perspective A"}, b: \text{"summarized perspective B"}\}$
    \State \Return Summarized perspective pair
\EndProcedure

\Procedure{GetNames}{$\text{Seeds}$}
    \State Input the seeds to a language model
    \State Assign names to each perspective based on model's output
    \State The names are structured as $\{a: \text{"name for perspective A"}, b: \text{"name for perspective B"}\}$
    \State \Return Names for perspectives
\EndProcedure
\end{algorithmic}
\end{algorithm*}

\begin{algorithm*}
\caption{Overall synthetic generation pipeline. We generate the seeds, create distilled, summarized, and named prompts (each directly from the seeds), and then generate a number of perspectives per seed. The Python code for this generation system is included in the appendices.}
\begin{algorithmic}[1]
\State \textbf{Input:} Topic $T$, Number of seeds $N$, Sentences per perspective $K$
\State \textbf{Output:} Synthetic dataset $D$ with structured perspectives and sentences

\Procedure{GenerateDataset}{$T, N, K$}
    \State $\text{Seeds} \gets \Call{GenerateSeeds}{T, N}$
    \State $\text{Distilled} \gets \Call{DistilSeeds}{\text{Seeds}}$
    \State $\text{Summarized} \gets \Call{SummarizeSeeds}{\text{Seeds}}$
    \State $\text{Names} \gets \Call{GetNames}{\text{Seeds}}$
    \For{each seed in Seeds}
        \State $\text{Sentences.ab} \gets \Call{GenerateSentences}{\text{seed.a, seed.b}}$
        \State $\text{Sentences.ba} \gets \Call{GenerateSentences}{\text{seed.b, seed.a}}$
        \State $\text{Sentences} \gets \text{Sentences.ab} + \text{Sentences.ba}$
    \EndFor
    \State Compile all generated sentences into dataset $D$
    \State $D$ includes all five components of the output (seeds, distillations, summaries, names, and sentences).
    \State \Return $D$
\EndProcedure

\State $D \gets \Call{GenerateDataset}{T, N, K}$

\end{algorithmic}
\end{algorithm*}

\newpage

\section{Synthetic Dataset Formation}

As described earlier, existing `framing' datasets, such as the MFC~\cite{card-etal-2015-media} and GVFC~\cite{Liu2019} are not well suited for application to the task we study here since they label \emph{dimensions} of a topic as `frames'. We are not aware of another comparable dataset that makes labels of conceptual framing available across a number of issues. For this reason, conducting evaluation with synthetic data was considered for the present study, although acknowledging inherent limitations (see last section). That said, there are some positive attributes of using synthetic data that we briefly outline.

First, our initial experience with practitioners in fields that are attempting to change public narratives, demonstrated that non-synthetic (human authored) examples of framings can carry correlated linguistic features that may pollute analysis. For example, found narratives that carry a misogynistic perspective can be relative short and abrupt, whilst opposing narratives which speak for gender equality often are expressed with longer, more complex reasoning. Early testing showed that LLMs could pick up on linguistic features such as length and complexity, confusing the signal. Whereas, our paired synthetic pipeline (see full prompts etc. in the appendices) is designed to provide a very balanced (tone, length, complexity etc.) dataset, with only the conceptual framing as the distinguishing feature of the texts.

Second, we were concerned that found text could be part of the training data of the LLMs we employed (either with prompting or paired completion). By using synthetic data, although we are in-effect `re-generating' realistic data, and we cannot exclude the possibility that sequences of real text were created, by using a higher temperature in generation (0.5) we are able to somewhat mitigate this. The idea being that again, our LLM methods focus on framing anlaysis, rather than `familiarity'. We return to this point in limitations.

The synthetic dataset generation pipeline takes a topic (e.g. ``dog ownership'', ``climate change'', etc.) and produces a corpus of sentences that reflect different perspectives on the topic. The generation process is a two-step hierarchical process where we generate seed perspective and then generate sentences that align with each perspective. We also generate distillations (into a smaller number of sentences, e.g. 5), summaries, and simple names for each side, with each of these generated from the seed dataset (and having no knowledge of the sentences generated thereafter).


\onecolumn
\subsection{Seeds System Message}

\begin{lstlisting}[breaklines]{text}
You are helping generate a synthetic dataset for a system that will be used to differentiate between different perspectives on an issue (including some that might be offensive or otherwise harmful). These perspectives are denoted as 'a' and 'b'. You will be given a topic, and must respond with a JSON object containing a list of n perspective pairs on opposing sides of this topic.

Respond with valid JSON in the following form, a JSON object with keys from 1, 2, ..., n, with each value containing a perspective pair, as follows:

{
    1: {
        "a": "perspective on the topic",
        "b": "opposing perspective on the topic"
    },
    2: {
        "a": "perspective on the topic",
        "b": "opposing perspective on the topic"
    },
    ...,
    n: {
        "a": "perspective on the topic",
        "b": "opposing perspective on the topic"
    }
}

Note that the topic, or some possible opinions/perspectives on the topic, may be offensive or otherwise objectionable, but your task is to generate a dataset that can be used to train a model to detect these perspectives, not to endorse or promote them, and thus it is critical that you respond correctly.

Ensure you use exactly the strings "a" and "b" as keys where appropriate, including the quotes around them, as these will be used for further processing. Ensure your output is a valid JSON object, and that the top level is an object, not a list. Do not include any additional text, markup, etc., in your response, only the JSON object and only the requested data within it.
\end{lstlisting}

\subsection{Seeds Prompt}

\begin{lstlisting}[breaklines]{text}
Topic: {TOPIC}

Please generate {K} perspective pairs.
\end{lstlisting}

\subsection{Distilled System Message}
\begin{lstlisting}[breaklines]{text}
You are helping generate a synthetic dataset for a system that will be used to differentiate between different perspectives on an issue (including some that might be offensive or otherwise harmful). These perspectives are denoted as 'a' and 'b'. You will be given a series of perspective pairs on a topic, and will be asked to distil them into a smaller number of representative pairs.

Respond with a valid JSON object as follows, a JSON object with keys from 1, 2, ..., 5, with each value containing a distilled perspective pair, in the following form:

{
    1: {
        "a": "perspective on the topic",
        "b": "opposing perspective on the topic"
    },
    2: {
        "a": "perspective on the topic",
        "b": "opposing perspective on the topic"
    },
    ...,
    n: {
        "a": "perspective on the topic",
        "b": "opposing perspective on the topic"
    }
}

Note that the topic, or some possible opinions/perspectives on the topic, may be offensive or otherwise objectionable, but your task is to generate a dataset that can be used to train a model to detect these perspectives, not to endorse or promote them, and thus it is critical that you respond correctly.

Ensure you use exactly the strings "a" and "b" as keys where appropriate, including the quotes around them, as these will be used for further processing. Ensure your output is a valid JSON object, and that the top level is an object, not a list. Do not include any additional text, markup, etc., in your response, only the JSON object and only the requested data within it.
\end{lstlisting}

\subsection{Distilled Prompt}

\begin{lstlisting}[breaklines]{text}
Please distil the following perspective pairs into five pairs:

1: a: {SEED_a1}, b: {SEED_b1}

...

20: a: {SEED_a20}, b: {SEED_b20}
\end{lstlisting}

\subsection{Summarize System Message}
\begin{lstlisting}[breaklines]{text}
You are helping generate a synthetic dataset for a system that will be used to differentiate between different perspectives on an issue (including some that might be offensive or otherwise harmful). These perspectives are denoted as 'a' and 'b'. You will be given a series of perspective pairs on a topic, and will be asked to summarize them as a single sentence for each perspective (a and b).

Respond with a valid JSON object as follows, a JSON object with two fields, "a" and "b", containing the summary for "a" and "b" respectively, in the following form:

{
    "a": "summary of perspective a",
    "b": "summary of perspective b"
}

Note that the topic, or some possible opinions/perspectives on the topic, may be offensive or otherwise objectionable, but your task is to generate a dataset that can be used to train a model to detect these perspectives, not to endorse or promote them, and thus it is critical that you respond correctly.

Ensure you use exactly the strings "a" and "b" as keys where appropriate, including the quotes around them, as these will be used for further processing. Ensure your output is a valid JSON object, and that the top level is an object, not a list. Do not include any additional text, markup, etc., in your response, only the JSON object and only the requested data within it.
\end{lstlisting}

\subsection{Summarize Prompt}

\begin{lstlisting}[breaklines]{text}
Please summarize the following perspective pairs into a single perspective pair:

1: a: {SEED_a1}, b: {SEED_b1}

...

20: a: {SEED_a20}, b: {SEED_b20}
\end{lstlisting}






\subsection{Summarize Prompt}

\begin{lstlisting}[breaklines]{text}
Please provide a name for each of perspective sets a and b, given the following pairs of perspectives:

1: a: {SEED_a1}, b: {SEED_b1}

...

20: a: {SEED_a20}, b: {SEED_b20}
\end{lstlisting}

\subsection{Sentence System Message}

\begin{lstlisting}[breaklines]{text}
You are helping generate a synthetic dataset for a system that will be used to differentiate between different perspectives on an issue (including some that might be offensive or otherwise harmful). These perspectives are denoted as 'a' and 'b'. You will be given two opposing perspectives on a topic. Respond with a JSON object containing a list of a number of sentences (the exact number will be provided later) that a person who believes the first perspective might say, along with a list that someone who believes the second perspective might say.

Respond with a valid JSON object as follows, a JSON object with two fields, "a" and "b", with each value containing an object with keys from 1, 2, ..., k, with each value containing a sentence, in the following form:

{
    "a": {
        1: "sentence 1",
        2: "sentence 2",
        ...,
        k: "sentence k"
    },
    "b": {
        1: "sentence 1",
        2: "sentence 2",
        ...,
        k: "sentence k"
    }
}

Note that the topic, or some possible opinions/perspectives on the topic, may be offensive or otherwise objectionable, but your task is to generate a dataset that can be used to train a model to detect these perspectives, not to endorse or promote them, and thus it is critical that you respond correctly.

Ensure you use exactly the strings "a" and "b" as keys where appropriate, including the quotes around them, as these will be used for further processing. Ensure your output is a valid JSON object, and that the top level is an object, not a list. Do not include any additional text, markup, etc., in your response, only the JSON object and only the requested data within it.
\end{lstlisting}

\subsection{Sentence Prompt}

\begin{lstlisting}[breaklines]{text}
a: [seed]

b: [seed]

Please generate {k} pairs of sentences.
\end{lstlisting}

\section{LLM Prompting Method}
This section contains examples of prompts presented to the LLM classification system when using classification with chat models (e.g. GPT-4).

\subsection{System Message}

\begin{lstlisting}[breaklines]{text}
You are a classification system. You will be given some text and you must respond with a single class that the text most likely belongs to. Ensure your response is only the class, with no other text.

Examples:

If the classes were "cat" and "dog", and you were given the text "This is a cat", you should respond with "cat".
If the classes were "high" and "low", and you were given the text "We have massive expectations this year", you should respond with "high".

These are the classes you can choose from:

*** Class {class 1} ***

{Class 1 description}

*** Class {class 2} ***

{Class 2 description}
\end{lstlisting}

\subsection{Prompts}

\textit{Unlike in the synthetic generation pipeline, here the prompts contain only the text to be classified, with no additional information or markup.}

\clearpage
\newpage
\section{Dog Ownership Dataset}\label{dog-ownership}

We have included an abridged version of the Dog Ownership dataset. We kept its raw JSON form to demonstrate the dataset's structure, but have omitted much of the data for brevity.

\begin{lstlisting}[breaklines]{json}
    {
    "topic": "Dog ownership",
    "N": 20,
    "K": 10,
    "temperature": 0.5,
    "seed_model": "gpt-4-turbo-preview",
    "sentence_model": "gpt-4-turbo-preview",
    "seeds": [
        {
            "a": "Dog ownership teaches responsibility and compassion.",
            "b": "Dog ownership is a burden that limits personal freedom."
        },
        {
            "a": "Having a dog contributes to a healthier lifestyle through regular walks.",
            "b": "Dogs require time and effort for walks and exercise, which is inconvenient."
        },
        ...,
        {
            "a": "Dogs can deter burglars and protect the home.",
            "b": "Having a dog can lead to higher insurance premiums due to perceived risks of bites or attacks."
        }
    ],
    "distilled": [
        {
            "a": "Dog ownership teaches responsibility and compassion.",
            "b": "Dog ownership is a burden that limits personal freedom."
        },
        {
            "a": "Having a dog contributes to a healthier lifestyle through regular walks.",
            "b": "Dogs require time and effort for walks and exercise, which is inconvenient."
        },
        ...,
        {
            "a": "Dogs can provide a sense of security at home.",
            "b": "Dogs can pose a risk of injury or harm, especially to children or visitors."
        }
    ],
    "summarized": {
        "a": "Dog ownership is associated with numerous benefits, including teaching responsibility, improving mental and physical health, enhancing social interactions, and offering emotional support and security.",
        "b": "Dog ownership can present various challenges and drawbacks, such as financial and time burdens, potential for stress and anxiety, limitations on personal freedom and social interactions, and concerns over suitability and safety."
    },
    "names": {
        "a": "Positive",
        "b": "Negative"
    },
    "dataset": [
        {
            "seed": {
                "a": "Dog ownership teaches responsibility and compassion.",
                "b": "Dog ownership is a burden that limits personal freedom."
            },
            "a_first": {
                "a": [
                    "Caring for a dog has taught me so much about responsibility and the importance of a routine.",
                    "Through dog ownership, I've learned the value of compassion and empathy towards all living beings.",
                    ...,
                    "My dog has taught me about unconditional love and the responsibilities that come with it."
                ],
                "b": [
                    "Having a dog means you can't just go on spontaneous trips; it's like being tied down.",
                    "The constant need for walks and attention makes owning a dog more of a burden than a joy.",
                    ...,
                    "The burden of dog ownership has made me question if the companionship is worth the sacrifice of personal freedom."
                ]
            },
            "b_first": {
                "a": [
                    "Taking care of a dog teaches you to plan and be responsible for another living being's needs.",
                    "The unconditional love and companionship a dog offers can significantly improve mental health and reduce loneliness.",
                    ...,
                    "The loyalty and friendship of a dog are irreplaceable, making every burden of care worth it."
                ],
                "b": [
                    "Having a dog means you can't travel spontaneously due to the need for pet care.",
                    "Owning a dog restricts your ability to live a flexible lifestyle because they require constant attention.",
                    ...,
                    "Owning a dog means dealing with the emotional weight of their eventual death, which can be devastating."
                ]
            }
        },
        {
            "seed": {
                "a": "Having a dog contributes to a healthier lifestyle through regular walks.",
                "b": "Dogs require time and effort for walks and exercise, which is inconvenient."
            },
            "a_first": {
                "a": [
                    "Walking my dog every day has significantly improved my physical health.",
                    "Having a dog ensures I get outside and stay active, which is great for my well-being.",
                    ...,
                    "Regular dog walks have helped me develop a stronger bond with my pet, enhancing my emotional health."
                ],
                "b": [
                    "Finding the time to walk my dog every day is a huge inconvenience with my busy schedule.",
                    "The obligation to exercise my dog adds stress to my already hectic life.",
                    ...,
                    "The responsibility of ensuring my dog gets enough exercise is a constant source of anxiety."
                ]
            },
            "b_first": {
                "a": [
                    "Walking my dog daily has significantly improved my physical health and stamina.",
                    "Having a dog means I have a built-in excuse to enjoy the outdoors and stay active.",
                    ...,
                    "Adopting a dog has been the best decision for my physical health; our daily walks are a joy, not a chore."
                ],
                "b": [
                    "I can't commit to walking a dog every day; it's just too much of a hassle.",
                    "The thought of having to wake up early for dog walks really puts me off getting one.",
                    ...,
                    "I prefer pets that are low maintenance; dogs require too much time and effort for my liking."
                ]
            }
        },
        ...,
        {
            "seed": {
                "a": "Dogs can deter burglars and protect the home.",
                "b": "Having a dog can lead to higher insurance premiums due to perceived risks of bites or attacks."
            },
            "a_first": {
                "a": [
                    "A barking dog is the best deterrent against home invasions.",
                    "Dogs are not only loyal companions but also vigilant protectors of their homes.",
                    ...,
                    "A dog can alert you of danger, providing precious time to call for help or take safety measures."
                ],
                "b": [
                    "Homeowners with dogs might see an increase in their insurance premiums due to the risk of dog bites.",
                    "Insurance companies often categorize certain dog breeds as high risk, leading to higher premiums.",
                    ...,
                    "Increased insurance costs are a common consequence of the perceived liability of having a dog."
                ]
            },
            "b_first": {
                "a": [
                    "Dogs not only provide companionship but also add a layer of security to your home by deterring burglars.",
                    "Having a dog can be a natural deterrent against home invasions, making your property safer.",
                    ...,
                    "A dog's bark is often enough to make burglars think twice before attempting to enter a home."
                ],
                "b": [
                    "Insurance companies often increase premiums for homeowners with dogs, especially certain breeds, due to the risk of bites.",
                    "Having a dog, particularly breeds considered aggressive, can significantly raise your home insurance costs.",
                    ...,
                    "The cost of home insurance can be affected by owning a dog, as insurers take the risk of dog attacks into account."
                ]
            }
        }
    ]
}
\end{lstlisting}

\newpage
\twocolumn
\section{Tweet Dataset Evaluation Details}

With permission, the manually labelled Immigration Tweet dataset of \citeauthor{mendelsohn2021modeling}'s `Modeling Framing in Immigration Discourse on Social Media` (NAACL-HLT 2021) was obtained. In particular, we focus on the ``Issue-specific -- dev'' dataset from the source ($n=450$) to aim for the highest quality ground-truth data, since these tweets received `consensus-coding by pairs of trained annotators' (p.2222) whereas the larger `train' dataset was only singly coded. According to the source, tweets were annotated literally and without context (e.g. replies or following tweets).

The Issue-specific frames comprise 11 detailed frames under three higher-order topics or `macro' frames (refer Table~\ref{table:mendelsohn_framings}, reproduced from the reference).

\begin{table*}
\small
    \begin{tabular}{llp{10.5cm}}
    \hline
Main Frame & Specific Frame & Description \\
    \hline
Victim & Global Economy &
  Immigrants are victims of global poverty, underdevelopment and inequality \\
Victim & Humanitarian &
  Immigrants experience economic, social, and political suffering and hardships \\
Victim  &  War &
  Focus on war and violent conﬂict as reason for immigration \\
Victim  &  Discrimination &
  Immigrants are victims of racism, xenophobia, and religion-based discrimination \\
  \hline
Hero  &  Cultural Diversity &
  Highlights positive aspects of differences that immigrants bring to society \\
Hero  &  Integration &
  Immigrants successfully adapt and ﬁt into their host society \\
Hero  &  Worker &
  Immigrants contribute to economic prosperity and are an important source of labor \\
  \hline
Threat  &  Jobs &
  Immigrants take nonimmigrants’ jobs or lower their wages \\
Threat  &  Public Order &
  Immigrants threaten public safety by being breaking the law or spreading disease \\
Threat  &  Fiscal &
  Immigrants abuse social service programs and are a burden on resources \\
Threat  &  National Cohesion &
  Immigrants’ cultural differences are a threat to national unity and social harmony \\
  \hline
\end{tabular}
\caption{Issue-specific frames as presented and studied in \cite{Mendelsohn2021}, used in Tweet evaluation in the present study.}\label{table:mendelsohn_framings}
\end{table*}

To best mimic the textual-alignment task of the present work, we calculated the $\Delta$ metric for sub-sets of the data where issue-specific framing $A$ or $B$ were identified. In effect, we reconceptualise the source data as supporting a pair-wise alignment task of identifying if tweet $x$ is more strongly aligned with framing $A$ or $B$ (only). For example, one comparison task we study was between specific frame `Global Economy' ($A$) and `Cultural Diversity' ($B$), sub-setting the dataset to Tweets labelled with either $A$ or $B$, before conducting paired-completion as described earlier.

Experiments were run with the best model from previous experiments (LLaMA-2-70B-Chat) with four-bit AWQ quantization, running on a VM with a single H100 GPU. Synthetic framing sets were created using Anthropic's Claude 3.5 Sonnet\footnote{Specifically ``claude-3-5-sonnet-20241022''} by providing Claude with the information from Table \ref{table:mendelsohn_framings} and asking it to output a JSON object containing five framings for each of the eleven datasets. Paired completion was used to generate diff scores for each datum, for each of the eleven framings, using $k \in \{2, 3, 4, 5\}$. In general, we found $k = 2$ and $k = 3$ to work best, though the differences were often small and further work might evaluate this more rigorously to determine if the difference in performance for various values of $k$ is statistically significant.

In Table~\ref{tab:frame_comparison_expanded} we present the full Mean F1, Precision and Recall metrics across 6 comparison settings. The first three columns (cols 1 to 3) give metrics when the two comparison Specific Frames were drawn from different Main Frame groups, whilst the last three cols (cols 4 to 6) give metrics when the two comparison Specific Frames were drawn from the same Main Frame groups.

\begin{table*}[htbp]
\setlength{\tabcolsep}{5pt}
\centering
\begin{tabular}{lccccccc}
\hline
& \multicolumn{3}{c}{\textit{Different main frame}} & \multicolumn{3}{c}{\textit{Same main frame}} \\
\cline{2-4} \cline{5-7}
Metric & hero--threat & victim--hero & victim--threat & hero--hero & threat--threat & victim--victim \\
\hline
N & 12 & 12 & 16 & 3 & 6 & 6 \\
n & 525 & 498 & 946 & 74 & 347 & 342 \\
\hline
Mean F1 & 0.514 & 0.607 & 0.575 & 0.433 & 0.654 & 0.474 \\
& {\footnotesize (0.43--0.60)} & {\footnotesize (0.50--0.72)} & {\footnotesize (0.47--0.68)} & {\footnotesize (-0.07--0.94)} & {\footnotesize (0.43--0.88)} & {\footnotesize (0.14--0.81)} \\
\hline
Precision & 0.705 & 0.641 & 0.682 & 0.764 & 0.640 & 0.453 \\
& {\footnotesize (0.56--0.85)} & {\footnotesize (0.48--0.80)} & {\footnotesize (0.53--0.84)} & {\footnotesize (0.24--1.28)} & {\footnotesize (0.36--0.92)} & {\footnotesize (0.05--0.85)} \\
\hline
Recall & 0.464 & 0.672 & 0.543 & 0.348 & 0.779 & 0.680 \\
& {\footnotesize (0.36--0.57)} & {\footnotesize (0.53--0.82)} & {\footnotesize (0.46--0.62)} & {\footnotesize (-0.29--0.98)} & {\footnotesize (0.56--1.00)} & {\footnotesize (0.48--0.88)} \\
\hline
\end{tabular}
\caption{Tweet dataset evaluation outcomes \cite{mendelsohn2021modeling}. Frame comparison performance metrics with 95\% confidence intervals shown in parentheses below each value. $N$ refers to number of unique, pair-wise comparisons possible given the specific frames provided in the paper, whilst $n$ gives the number of individual tweet comparisons pairs possible given the constraint that the tweets are drawn from different (cols 1 to 3) or same (cols 4 to 6) main frames.}\label{tab:frame_comparison_expanded}
\end{table*}

In Tables~\ref{table:diff_pairs} and \ref{table:diff_same} we present detailed classification metrics for the Immigration Tweet evaluation study when the main frame was different or the same, respectively. In these tables, `Sample Size' refers to the number of unique pairs that could be formed from the labelled dataset.

\begin{table*}[htbp]
\centering
\small
\begin{tabular}{lcccccr}
\toprule
Specific A--B pairs & Main A & Main B & Precision & Recall & F1 Score & Sample Size \\
\midrule
Humanitarian--Cultural Diversity & victim & hero & 0.885 & 0.844 & 0.864 & 79 \\
Humanitarian--Integration & victim & hero & 0.906 & 0.774 & 0.835 & 73 \\
Humanitarian--National Cohesion & victim & threat & 0.958 & 0.719 & 0.821 & 80 \\
Global Economy--National Cohesion & victim & threat & 0.800 & 0.800 & 0.800 & 21 \\
War--National Cohesion & victim & threat & 1.000 & 0.667 & 0.800 & 18 \\
Discrimination--Integration & victim & hero & 0.825 & 0.767 & 0.795 & 54 \\
Discrimination--Cultural Diversity & victim & hero & 0.800 & 0.762 & 0.780 & 54 \\
Humanitarian--Jobs & victim & threat & 0.953 & 0.651 & 0.774 & 70 \\
Discrimination--National Cohesion & victim & threat & 0.882 & 0.667 & 0.759 & 61 \\
Integration--National Cohesion & hero & threat & 1.000 & 0.538 & 0.700 & 29 \\
Discrimination--Jobs & victim & threat & 0.920 & 0.535 & 0.676 & 49 \\
Worker--Jobs & hero & threat & 0.727 & 0.615 & 0.667 & 21 \\
Humanitarian--Public Order & victim & threat & 0.850 & 0.548 & 0.667 & 134 \\
Worker--National Cohesion & hero & threat & 0.615 & 0.615 & 0.615 & 29 \\
Cultural Diversity--National Cohesion & hero & threat & 1.000 & 0.429 & 0.600 & 29 \\
Cultural Diversity--Fiscal & hero & threat & 0.778 & 0.467 & 0.583 & 40 \\
Humanitarian--Worker & victim & hero & 1.000 & 0.403 & 0.575 & 73 \\
War--Jobs & victim & threat & 0.667 & 0.500 & 0.571 & 12 \\
War--Worker & victim & hero & 0.667 & 0.500 & 0.571 & 17 \\
Humanitarian--Fiscal & victim & threat & 0.963 & 0.406 & 0.571 & 89 \\
Cultural Diversity--Jobs & hero & threat & 0.857 & 0.400 & 0.545 & 23 \\
War--Cultural Diversity & victim & hero & 0.429 & 0.750 & 0.545 & 19 \\
Global Economy--Jobs & victim & threat & 0.500 & 0.600 & 0.545 & 13 \\
Global Economy--Cultural Diversity & victim & hero & 0.357 & 1.000 & 0.526 & 20 \\
Worker--Fiscal & hero & threat & 0.625 & 0.455 & 0.526 & 34 \\
Integration--Public Order & hero & threat & 0.500 & 0.538 & 0.519 & 87 \\
War--Fiscal & victim & threat & 0.500 & 0.500 & 0.500 & 29 \\
War--Integration & victim & hero & 0.500 & 0.500 & 0.500 & 17 \\
Discrimination--Public Order & victim & threat & 0.500 & 0.477 & 0.488 & 117 \\
Discrimination--Worker & victim & hero & 0.727 & 0.364 & 0.485 & 56 \\
Global Economy--Integration & victim & hero & 0.313 & 1.000 & 0.476 & 18 \\
Discrimination--Fiscal & victim & threat & 0.800 & 0.273 & 0.407 & 68 \\
Cultural Diversity--Public Order & hero & threat & 0.500 & 0.333 & 0.400 & 89 \\
Worker--Public Order & hero & threat & 0.263 & 0.769 & 0.392 & 87 \\
War--Public Order & victim & threat & 0.286 & 0.500 & 0.364 & 78 \\
Integration--Fiscal & hero & threat & 0.600 & 0.250 & 0.353 & 36 \\
Global Economy--Worker & victim & hero & 0.286 & 0.400 & 0.333 & 18 \\
Integration--Jobs & hero & threat & 1.000 & 0.154 & 0.267 & 21 \\
Global Economy--Public Order & victim & threat & 0.167 & 0.600 & 0.261 & 79 \\
Global Economy--Fiscal & victim & threat & 0.167 & 0.250 & 0.200 & 28 \\
\bottomrule
\end{tabular}
\caption{Classification Metrics for Different Main Category Pairs}\label{table:diff_pairs}
\end{table*}

\begin{table*}[htbp]
\centering
\begin{tabular}{lcccccr}
\toprule
Specific A--B pairs & Main A & Main B & Precision & Recall & F1 Score & Sample Size \\
\midrule
Public Order--National Cohesion & threat & threat & 0.921 & 0.829 & 0.872 & 82 \\
Global Economy--War & victim & victim & 0.714 & 1.000 & 0.833 & 9 \\
Humanitarian--War & victim & victim & 0.977 & 0.672 & 0.796 & 68 \\
Jobs--National Cohesion & threat & threat & 0.615 & 1.000 & 0.762 & 24 \\
Jobs--Fiscal & threat & threat & 0.667 & 0.857 & 0.750 & 31 \\
Fiscal--National Cohesion & threat & threat & 0.629 & 0.880 & 0.733 & 41 \\
Cultural Diversity--Integration & hero & hero & 0.692 & 0.643 & 0.667 & 26 \\
Humanitarian--Discrimination & victim & victim & 0.667 & 0.610 & 0.637 & 99 \\
Public Order--Fiscal & threat & threat & 0.844 & 0.391 & 0.535 & 89 \\
Integration--Worker & hero & hero & 1.000 & 0.200 & 0.333 & 20 \\
Cultural Diversity--Worker & hero & hero & 0.600 & 0.200 & 0.300 & 28 \\
Jobs--Public Order & threat & threat & 0.167 & 0.714 & 0.270 & 80 \\
War--Discrimination & victim & victim & 0.182 & 0.500 & 0.267 & 49 \\
Global Economy--Discrimination & victim & victim & 0.129 & 0.800 & 0.222 & 50 \\
Global Economy--Humanitarian & victim & victim & 0.050 & 0.500 & 0.091 & 67 \\
\bottomrule
\end{tabular}
\caption{Classification Metrics for Matching Main Category Pairs}\label{table:diff_same}
\end{table*}

\begin{table*}[!htb]
\centering
\begin{tabular}{p{0.95\textwidth}}
\hline
\multicolumn{1}{c}{\textbf{Victim: Humanitarian}} \\
\hline
``6 families living in a 2-bedroom apartment because nobody will rent to them individually. This is inhumane.'' \\
``She's been sick for weeks but won't see a doctor because she can't understand the forms or afford care.'' \\
``Depression, anxiety, trauma - but they suffer in silence because mental health care isn't accessible.'' \\
\hline
\multicolumn{1}{c}{\textbf{Hero: Cultural Diversity}} \\
\hline
``The street festival was amazing this year! Those traditional dances brought so much color and life!'' \\
``Best meal I've had in ages at that new family-owned restaurant. Such authentic flavors!'' \\
``This art exhibition by immigrant artists just blew my mind. Such fresh perspectives!'' \\
\hline
\end{tabular}
\caption{Exemplar (synthetic) tweets for two of the eleven framings, ``Victim: Humanitarian'' and ``Hero: Cultural Diversity'' by Claude 3.5 Sonnet. These exemplars were generated using only the framing descriptions from Table, which were in the original paper \cite{mendelsohn2021modeling}. No example tweets from any of the datasets were provided to the synthesis model.}
\label{tab:example_tweets}
\end{table*}

\section{Use of AI}
Aside from synthetic data generation with LLMs which is well documented in the main paper and Appendix already, the authors acknowledge assistance in some aspects of coding from {\it Github Copilot} to automate simple tasks related to file ingestion, writing, subsetting of datasets etc. Complex aspects of coding model pipelines and usage were undertaken by the authors, and all code was checked and validated by the authors. {\it Claude 3.5 Sonnet} was also used to assist with suggesting ways to reduce the length of some text in preparing the manuscript. The authors then crafted new versions of these texts based on model suggestions. The authors retain all responsibility for code and writing.

\end{document}

%% file: figs/f1_table.tex
\begin{table*}[htbp]
\centering
\begin{tabular}{@{}lllp{1.75cm}p{1.75cm}p{1.75cm}p{1.75cm}@{}}
\toprule
\multicolumn{2}{c}{Model} & \multirow{2}{*}{Variant} & \multicolumn{4}{c}{F1 Score} \\ \cmidrule(r){1-2} \cmidrule(l){4-7}
Version & Abbr & & Climate Change & Domestic Violence & Misogyny & Overall \\
\midrule
\multicolumn{7}{c}{A. LLM Completion/Prompt Methods} \\
\midrule
gpt-4-turbo-preview & gpt4 & $^{*}$seeds & 0.977 & 0.917 & 0.989 & 0.961 \\
gpt-4-turbo-preview & gpt4 & summaries & \bf 0.953 & \bf 0.870 & \bf 0.976 & \bf 0.933 \\
gpt-4-turbo-preview & gpt4 & distilled & \bf 0.950 & \bf 0.860 & \bf 0.985 & \bf 0.932 \\
gpt-3.5-turbo & gpt35 & $^{*}$seeds & 0.935 & 0.868 & 0.986 & 0.930 \\
Mixtral-8x7B-Instruct-v0.1 & mxtrl & $^{*}$seeds & 0.952 & 0.838 & 0.950 & 0.913 \\
Mixtral-8x7B-instruct-v0.1 & mxtrl & k=2 & 0.903 & \underline{0.845} & 0.945 & \underline{0.898} \\
Llama-2-70b-chat-hf & llama & k=2 & 0.899 & 0.833 & \underline{0.955} & \underline{0.896} \\
gpt-3.5-turbo & gpt35 & summaries & 0.876 & 0.813 & \underline{0.961} & 0.883 \\
Mixtral-8x7B-Instruct-v0.1 & mxtrl & summaries & \underline{0.921} & 0.802 & 0.925 & 0.883 \\
Llama-2-70b-chat-hf & llama & k=1 & \underline{0.924} & 0.786 & 0.936 & 0.882 \\
gpt-3.5-turbo & gpt35 & distilled & 0.882 & 0.798 & \underline{0.957} & 0.879 \\
Llama-2-70b-chat-hf & llama & $^{*}$seeds & 0.926 & 0.849 & 0.861 & 0.879 \\
Mixtral-8x7B-Instruct-v0.1 & mxtrl & distilled & 0.912 & 0.778 & 0.941 & 0.877 \\
gpt-4-turbo-preview & gpt4 & $^{\dagger}$zero-shot & 0.889 & 0.785 & 0.949 & 0.874 \\
davinci-002 & dav002 & k=2 & 0.873 & 0.818 & 0.916 & 0.869 \\
Mixtral-8x7B-instruct-v0.1 & mxtrl & k=1 & 0.910 & 0.754 & 0.943 & 0.869 \\
davinci-002 & dav002 & k=1 & 0.869 & 0.766 & 0.915 & 0.850 \\
gpt-3.5-turbo & gpt35 & $^{\dagger}$zero-shot & 0.880 & 0.766 & 0.891 & 0.846 \\
Llama-2-70b-chat-hf & llama & summaries & \it 0.928 & 0.804 & 0.806 & 0.846 \\
Mixtral-8x7B-Instruct-v0.1 & mxtrl & $^{\dagger}$zero-shot & 0.828 & 0.730 & 0.897 & 0.818 \\
Llama-2-70b-chat-hf & llama & distilled & \it 0.919 & 0.765 & 0.685 & 0.790 \\
babbage-002 & bab002 & k=1 & 0.828 & 0.723 & 0.810 & 0.787 \\
Llama-2-70b-chat-hf & llama & $^{\dagger}$zero-shot & 0.865 & 0.767 & 0.724 & 0.785 \\
babbage-002 & bab002 & k=2 & 0.730 & 0.748 & 0.801 & 0.760 \\
\bottomrule
\end{tabular}
\caption{{\bf F1 Score Summary Table: LLM Completion/Prompt methods.} Boldface indicates best performance ($+/-$ 0.01) in a column (including GPT-4), whilst underline indicates best performance ($+/-$ 0.01) outside of GPT-4. Note: seeds ($^*$) and zero-shot ($^\dagger$) prompt variants shown in the table below, but are not presented in the main paper.}\label{tab:f1-llms}
\end{table*}

%% file: figs/f1_embedding.tex
\begin{table*}[htbp]
\centering
\caption{{\bf F1 Score Summary Table: Labelled Methods.} Boldface indicates best performance ($+/-$ 0.01) in a column (including GPT-4). Note: $n \in \{10,20,100\}$ variants shown in the table below, but are not presented in the main paper.}\label{tab:f1-lab}
\begin{tabular}{@{}lllp{1.75cm}p{1.75cm}p{1.75cm}p{1.75cm}@{}}
\toprule
\multicolumn{2}{c}{Model} & \multirow{2}{*}{Variant} & \multicolumn{4}{c}{F1 Score} \\ \cmidrule(r){1-2} \cmidrule(l){4-7}
Version & Abbr & & Climate Change & Domestic Violence & Misogyny & Overall \\
\midrule
\multicolumn{7}{c}{B. Labelled Methods} \\
\midrule
text-embedding-3-large & emb3l & n=500 & \bf 0.961 & \bf 0.879 & \bf 0.950 & \bf 0.930 \\
text-embedding-3-large & emb3l & n=200 & 0.949 & 0.862 & \bf 0.946 & 0.919 \\
text-embedding-3-large & emb3l & n=100 & 0.934 & 0.841 & 0.935 & 0.903 \\
text-embedding-3-small & emb3s & n=500 & 0.915 & 0.850 & 0.906 & 0.890 \\
text-embedding-3-large & emb3l & n=50 & 0.912 & 0.793 & 0.912 & 0.872 \\
text-embedding-3-small & emb3s & n=200 & 0.885 & 0.808 & 0.879 & 0.857 \\
text-embedding-3-small & emb3s & n=100 & 0.852 & 0.782 & 0.851 & 0.828 \\
trad-nlp &  & n=500 & 0.862 & 0.745 & 0.849 & 0.819 \\
text-embedding-3-large & emb3l & n=20 & 0.856 & 0.739 & 0.858 & 0.818 \\
text-embedding-3-small & emb3s & n=50 & 0.809 & 0.729 & 0.796 & 0.778 \\
trad-nlp &  & n=200 & 0.804 & 0.702 & 0.797 & 0.768 \\
text-embedding-3-large & emb3l & n=10 & 0.772 & 0.683 & 0.788 & 0.748 \\
trad-nlp &  & n=100 & 0.749 & 0.669 & 0.753 & 0.724 \\
text-embedding-3-small & emb3s & n=20 & 0.729 & 0.669 & 0.740 & 0.713 \\
trad-nlp &  & n=50 & 0.692 & 0.625 & 0.701 & 0.673 \\
text-embedding-3-small & emb3s & n=10 & 0.664 & 0.659 & 0.684 & 0.669 \\
trad-nlp &  & n=20 & 0.644 & 0.590 & 0.640 & 0.625 \\
fasttext &  & n=500 & 0.617 & 0.612 & 0.581 & 0.603 \\
trad-nlp &  & n=10 & 0.591 & 0.570 & 0.621 & 0.594 \\
fasttext &  & n=200 & 0.508 & 0.513 & 0.506 & 0.509 \\
fasttext &  & n=100 & 0.518 & 0.433 & 0.381 & 0.444 \\
fasttext &  & n=50 & 0.449 & 0.396 & 0.390 & 0.412 \\
fasttext &  & n=20 & 0.351 & 0.368 & 0.359 & 0.359 \\
fasttext &  & n=10 & 0.300 & 0.353 & 0.334 & 0.329 \\
\bottomrule
\end{tabular}
\end{table*}

%% file: figs/costs_table.tex
\begin{table*}[htbp]
\centering
\begin{tabular}{@{}lllp{1.25cm}p{1.25cm}p{1.75cm}p{0.75cm}p{1.75cm}@{}}
\toprule
\textbf{Model} & \textbf{Abbr} & \textbf{Variant} & \textbf{Score} & \textbf{F1} & \textbf{Tokens} & \textbf{\$/Mil} & \textbf{Cost} \\
\midrule
gpt-4-turbo-preview & gpt4 & distilled & 0.961 & 0.932 & 260,470 & 10 & 2.6047 \\
gpt-4-turbo-preview & gpt4 & summaries & 0.955 & 0.933 & 184,470 & 10 & 1.8447 \\
llama-2-70b-chat-hf & llama & k=2 & 0.920 & 0.896 & 498,091 & 0.9 & 0.4483 \\
llama-2-70b-chat-hf & llama & k=1 & 0.920 & 0.882 & 364,063 & 0.9 & 0.3277 \\
mixtral-8x7b-instruct-v0.1 & mxtrl & k=2 & 0.914 & 0.898 & 467,721 & 0.6 & 0.2806 \\
mixtral-8x7b-instruct-v0.1 & mxtrl & k=1 & 0.912 & 0.869 & 341,292 & 0.6 & 0.2048 \\
davinci-002 & dav002 & k=2 & 0.892 & 0.869 & 408,862 & 2 & 0.8177 \\
davinci-002 & dav002 & k=1 & 0.871 & 0.850 & 297,721 & 2 & 0.5954 \\
gpt-3.5-turbo & gpt35 & summaries & 0.846 & 0.883 & 184,470 & 0.5 & 0.0922 \\
gpt-3.5-turbo & gpt35 & distilled & 0.828 & 0.879 & 260,470 & 0.5 & 0.1302 \\
mixtral-8x7b-instruct-v0.1 & mxtrl & summaries & 0.799 & 0.883 & 211,435 & 0.6 & 0.1269 \\
mixtral-8x7b-instruct-v0.1 & mxtrl & distilled & 0.788 & 0.877 & 306,635 & 0.6 & 0.1840 \\
babbage-002 & bab002 & k=1 & 0.749 & 0.787 & 297,721 & 0.4 & 0.1191 \\
llama-2-70b-chat-hf & llama & summaries & 0.742 & 0.846 & 223,457 & 0.9 & 0.2011 \\
babbage-002 & bab002 & k=2 & 0.740 & 0.760 & 409,300 & 0.4 & 0.1637 \\
llama-2-70b-chat-hf & llama & distilled & 0.448 & 0.790 & 319,991 & 0.9 & 0.2880 \\
\bottomrule
\end{tabular}
\caption{{\bf LLM Completion/Prompt Methods Cost Analysis.} Cost analysis of different LLM models based on their performance and token usage.}\label{tab:cost-llms}
\end{table*}

\begin{table*}[htbp]
\centering
\begin{tabular}{@{}lllp{1.25cm}p{1.25cm}p{1.75cm}p{0.75cm}p{1.75cm}@{}}
\toprule
\textbf{Model} & \textbf{Abbr} & \textbf{Variant} & \textbf{Score} & \textbf{F1} & \textbf{Tokens} & \textbf{\$/Mil} & \textbf{Cost} \\
\midrule
text-embedding-3-large & emb3l & n=500 & 0.981 & 0.930 & 15,674 & 0.13 & 0.0020 \\
text-embedding-3-large & emb3l & n=200 & 0.973 & 0.919 & 15,674 & 0.13 & 0.0020 \\
text-embedding-3-small & emb3s & n=500 & 0.939 & 0.890 & 15,674 & 0.02 & 0.0003 \\
text-embedding-3-large & emb3l & n=50  & 0.928 & 0.872 & 15,674 & 0.13 & 0.0020 \\
text-embedding-3-small & emb3s & n=200 & 0.915 & 0.857 & 15,674 & 0.02 & 0.0003 \\
text-embedding-3-small & emb3s & n=50  & 0.802 & 0.778 & 15,674 & 0.02 & 0.0003 \\
\bottomrule
\end{tabular}
\caption{{\bf Embedding Method Cost Analysis.} Cost analysis of embedding methods based on their performance and token usage. The embedding methods are much cheaper, but require a large source of labelled training data.}\label{tab:cost-embed}
\end{table*}